\documentclass{article}

\usepackage[preprint]{corl_2024}

\usepackage{graphicx} 
\usepackage{amsmath}
\usepackage{amsthm}
\usepackage{caption}
\usepackage{subcaption}
\usepackage{float}
\usepackage{algorithm}

\usepackage{algpseudocode}

\usepackage{amssymb}  
\usepackage{amsfonts}  

\newcommand{\E}{\mathop{\mathbb{E}}}

\title{Constrained Meta Agnostic Reinforcement \\Learning}

\author{
    Karam Daaboul\thanks{These authors contributed equally to this work.} \\
    daaboul@kit.edu \\
    KIT Karlsruhe Institute of Technology
    \And
    Florian Kuhm\footnotemark[1] \\
    florian.kuhm@student.kit.edu \\
    KIT Karlsruhe Institute of Technology
    \And
    Tim Joseph \\
    joseph@fzi.de \\
    FZI Research Center for Information Technology
    \And
    Marius J. Zöllner \\
    marius.zoellner@kit.edu\\
    KIT Karlsruhe Institute of Technology
}

\begin{document}
\maketitle

\begin{abstract}
Meta-Reinforcement Learning (Meta-RL) aims to acquire meta-knowledge for quick adaptation to diverse tasks. However, applying these policies in real-world environments presents a significant challenge in balancing rapid adaptability with adherence to environmental constraints. Our novel approach, Constraint Model Agnostic Meta Learning (C-MAML), merges meta learning with constrained optimization to address this challenge. C-MAML enables rapid and efficient task adaptation by incorporating task-specific constraints directly into its meta-algorithm framework during the training phase. This fusion results in safer initial parameters for learning new tasks. We demonstrate the effectiveness of C-MAML in simulated locomotion with mobile robot tasks of varying complexity, highlighting its practicality and robustness in different environments.
\end{abstract}



\section{Introduction}
\label{sec:Introduction}
Since its inception, Machine Learning, and particularly Reinforcement Learning (RL), have strived to emulate human learning processes in machines. A notable early example is Rosenblatt's development of the Perceptron in 1957, modeled after the human brain \citep{rosenblatt1957perceptron}. This approach continued to evolve, with RL drawing inspiration from the biological and psychological parallels observed in human and animal behavior \citep{sutton2018reinforcement}. A key aspect of human learning is the reliance on prior knowledge to acquire new skills efficiently. Unlike machines trained through classical RL, which typically starts from random initializations, humans leverage their accumulated experiences. This is exemplified in scenarios such as learning to drive a car, where prior knowledge of the dynamics of a car greatly accelerates the learning process. Addressing this challenge, the field of meta learning has emerged, producing a diverse array of algorithms aimed at mimicking this human-like efficiency in learning \citep{Finn17MAML, Nichol18Reptile, Rajeswaran19iMAML}. The objective is to train RL agents to adapt rapidly to various tasks with minimal new task-specific experiences. This adaptability was illustrated by Beck et al. (\citeyear{MetaRLSurvey23}) with the example of a cooking robot that, during Meta-Training, learns to navigate different kitchen layouts and appliance arrangements. Upon deployment in a customer's kitchen, the robot adjusts to this new environment with relative ease.

However, a critical issue largely overlooked in meta learning literature is the safety of the meta learning process itself. The agent should not behave in a manner that endangers itself or its surroundings, neither during the meta learning phase nor in subsequent deployment. For instance, a kitchen robot must be able to recognize the presence of people to prevent injuring them, even while adapting to an unfamiliar kitchen layout. While there is substantial safety-oriented research in RL \citep{Schulman15TRPO, Achiam17CPO, SafetyGymOpenAI, Zanger2021}, these studies typically focus on adhering to constraints during the training of an agent for a specific task. In contrast, this work presents a method that ensures safety during meta-training, generates a safe set of initial parameters and adheres to constraints during fine-tuning. The combination of safety and meta learning presents distinct challenges that have not been fully addressed in existing literature. The presented research introduces Constrained MAML (C-MAML), an innovative approach designed to improve safety during training and adaptation phases in meta-learning. This enhancement is achieved by incorporating constraint-based methods into the traditional Model Agnostic Meta Learning (MAML) framework. Additionally, we have expanded the C-MAML framework by developing a practical algorithm that incorporates first-order meta-gradient methods along with a global safety critic in the outer loop. This enhancement is designed to ensure computational efficiency and the safety of the learned meta-policy during its application. Furthermore, the results demonstrate the adaptability of C-MAML, highlighting its capability to function effectively with various safe RL methods in the inner loop. The paper's structure is as follows: Section \ref{sec:RelatedWorks} presents the state-of-the-art solutions in meta learning and safe RL, forming the foundation for our newly developed algorithm. Section \ref{sec:Preliminaries} introduces the essential preliminaries and notation. Our novel approach, C-MAML, is detailed in Chapter \ref{sec:Approach}, followed by experiments and an outlook in sections \ref{sec:Evaluation} and \ref{sec:Conclusion}.
\section{Related Works}
\label{sec:RelatedWorks}
\subsection{Meta Learning}
Meta-learning, or "learning to learn," aims at discovering a meta-initialization to enhance the speed of adaptation to new tasks, moving beyond traditional random initialization approaches \citep{MetaLearningSurvey}. In Meta-RL, this concept is extended to train a meta-policy across varied environments to facilitate quick adjustment to new challenges \citep{Lee2018}. Early Meta Learning efforts, like those by Chalmers et al. (\citeyear{CHALMERS1991}), utilized black-box strategies to evolve update rules for neural net weights. Meta-RL's black-box methods, such as $\textnormal{RL}^2$ \citep{Duan16RL2}, embed the learning algorithm within an RNN, while gradient-based strategies like MAML prioritize swift adaptation via gradient updates. Finn et al.'s (\citeyear{Finn17MAML}) MAML optimizes a model's initial parameters for improved task performance through a dual-level optimization process, beneficial for both model-free \citep{NEURIPS2018_4de75424} and model-based RL \citep{Daaboul2022}. FoMAML, a variant by Finn et al. (\citeyear{Finn17MAML}), simplifies MAML by omitting second-order gradients, offering efficiency with minimal performance loss. Similarly, REPTILE by Nichol et al. (\citeyear{Nichol18Reptile}) updates meta-parameters without second-order derivatives. iMAML \citep{Rajeswaran19iMAML} addresses information loss by indirectly incorporating the Jacobi matrix in the meta-gradient calculation. While these methodologies excel in fast adaptability, they typically overlook safety considerations in learning initialization.
\subsection{Safe Reinforcement Learning}
Research in safe RL focuses on ensuring adherence to safety constraints through techniques like gradient or parameter projection, exemplified by Projection-Based Constrained Policy Optimization (PBCPO) \citep{Yang202} and related methods. Dalal et al. (\citeyear{dalal2018}) introduced a strategy employing safety layers to adjust actions per state, bypassing the need for domain-specific action modifications. Concurrently, Chow et al. (\citeyear{Chow2018}) explored using Linear Programming to derive Lyapunov functions for enforcing agent safety constraints. Lagrangian duality methods, addressing optimization while respecting constraints are pivotal. This category includes Trust Region Policy Optimization (TRPO) and Proximal Policy Optimization (PPO), leveraging Lagrangian techniques \citep{SafetyGymOpenAI}. Achiam et al. (\citeyear{Achiam17CPO}) introduced Constrained Policy Optimization (CPO), a sophisticated second-order algorithm for identifying feasible policies within a trust region, ensuring both performance improvement and constraint adherence. Although these algorithms provide solutions for real-world RL applications, it is critical to have agents that can quickly adapt to new tasks in dynamic environments. This aspect is often overlooked in the pursuit of safety but is essential for practical deployment.
\vspace{-2mm} 
\section{Preliminaries}
\label{sec:Preliminaries}
Meta-RL aims to find a meta-policy that can efficiently adapt to a wide range of tasks from a task distribution $T$, aiming to maximize the expected return across these tasks. This is formalized as:
\begin{equation}
\max_{\pi}~\E_{p\sim T}\left(J_p(\pi_{p})\right).
\end{equation}
Here, $\pi$ denotes the meta-policy, which is adapted for each task $p$ in $T$ to a task-specific policy $\pi_{p}$, crucial for optimal performance across tasks. The adaptation of $\pi$ to $\pi_{p}$ for task $p$ is achieved through an update mechanism $U_p$:
\begin{equation}
\pi_{p} = U_p(\pi).
\end{equation}
In practice, $U_p$ might utilize an algorithm like gradient descent \citep{Finn17MAML}, updating the meta-policy based on the gradient of $J_p(\pi)$:
\begin{equation}
\pi_{p} = \pi - \alpha \cdot \nabla_{\pi} J_p(\pi).
\end{equation}
In the distribution \(T\), tasks are modeled as unique Constrained Markov Decision Processes (CMDPs) represented by the tuple \((\mathcal{S}, \mathcal{A}, r_p, \mathcal{C}_p, M, \mu, D_p)\). Here, \(\mathcal{S}\) and \(\mathcal{A}\) signify the state and action spaces, respectively. The reward function specific to task \(p\) is \(r_p: \mathcal{S} \times \mathcal{A} \times \mathcal{S} \rightarrow \mathbb{R}\), while \(\mathcal{C}_p\) consists of safety-related cost functions, with each \(c_{p,j}: \mathcal{S} \times \mathcal{A} \times \mathcal{S} \rightarrow \mathbb{R}\). The CMDP framework also includes the state transition probability \(M: \mathcal{S} \times \mathcal{A} \times \mathcal{S} \rightarrow [0,1]\) and the initial state distribution \(\mu: \mathcal{S} \rightarrow [0,1]\). Furthermore, \(D_p\) details the predefined cost limits \(\{d_{p,j} \in \mathbb{R}\}\) for each cost function \(c_{p,j}\) in task \(p\). Although all tasks utilize the same \(M\), they are distinguished by their specific reward and cost structures. The expected discounted returns and costs under a policy \(\pi(a|s)\) for task \(p\) are given by:
\[
J_p(\pi) = \mathbb{E}_{\tau \sim \pi}\left[\sum_{t=0}^{\infty} \gamma^t r_p(s_t, a_t, s_{t+1})\right], \quad J_{c_{p,j}}(\pi) = \mathbb{E}_{\tau \sim \pi}\left[\sum_{t=0}^{\infty} \gamma^t c_{p,j}(s_t, a_t, s_{t+1})\right].
\]
In this context, \(\tau\) represents trajectories starting from initial states \(s_0\), with actions \(a\) drawn according to \(\pi(a|s)\), and subsequent states resulting from \(M\). The discount factor is denoted by \(\gamma\). A policy \(\pi\) is safe if it meets the condition \(J_{c_{p,j}}(\pi) \leq d_{p,j}\) for every \(j\) in \(\mathcal{C}_p\), thus establishing a set of acceptable policies \(\Pi_{\mathcal{C}_p}\) for task \(p\). The aim of constrained policy optimization is thus  to optimize expected returns within these safety limits:
\[
\max_{\pi} J_p(\pi) \quad \text{s.t.} \quad J_{c_{p,j}}(\pi) \leq d_{p,j}, \forall j \in \mathcal{C}_p.
\]
 Given \( R_p(\tau) = \sum_{t=0}^\infty \gamma^t r_p(s_t, a_t, s_{t+1}) \) as the return for a trajectory \(\tau\), the value function \( V^\pi_p(s) \) and action-value function \( Q^\pi_p(s,a) \) are defined as the expected returns from state \(s\) and from taking action \(a\) in state \(s\) under policy \(\pi\), respectively:
\[
V^\pi_p(s) = \E_{\tau \sim (\pi)} [ R_p(\tau) | s_0 = s ], \quad Q^\pi_p(s, a) = \E_{\tau \sim (\pi)} [ R_p(\tau) | s_0 = s, a_0 = a ].
\]
The advantage function, \( A^\pi_p(s, a) = Q^\pi_p(s,a) - V^\pi_p(s) \), quantifies the relative benefit of action \(a\) in state \(s\) under \(\pi\). In addition to these functions, the CMDP setting also includes cost-related counterparts for each task \(p\) in \(T\), defining the total trajectory cost as \(C_{p,j}(\tau) = \sum_{t=0}^\infty \gamma^t \cdot c_{p,j}(s_t, a_t, s_{t+1})\). For expected cost assessments, it introduces \(V^\pi_{c_{p,j}}(s)\), \(Q^\pi_{c_{p,j}}(s, a)\), and \(A^\pi_{c_{p,j}}(s, a)\) corresponding to each cost function \(c_{p,j}\) in \(\mathcal{C}_p\). For simplicity, subsequent discussions assume each task has only one constraint.

Analysis often employs the stationary discounted state distribution \(d^\pi(s)\), the probability of visiting state \(s\) under policy \(\pi\) with transition probability \(P\). Kakade and Langford (\citeyear{KakadeLangford02}) describe the expected return difference between policies \(\pi\) and \(\pi'\) as:
\begin{align}\label{KakadeLangford}
    J({\pi'}) = J(\pi) + \E_{\substack{s \sim d^{{\pi'}} \\ a \sim \pi'}}\left(\sum_{t = 0}^{\infty} \gamma^t\cdot A^{\pi}(s,a)\right).
\end{align}
This captures expected returns' variation, combining state transitions and policy advantages. However, direct sampling for optimization is intractable. Schulman et al.\ (\citeyear{Schulman15TRPO}) used importance sampling to address this:
\begin{align}
    J({\pi'}) = J(\pi) + \E_{\substack{s \sim d^{{\pi'}} \\ a \sim \pi}}\left(\sum_{t = 0}^{\infty} \gamma^t\cdot \frac{\pi'(a|s)}{\pi(a|s)}\cdot A^{\pi}(s,a)\right).
\end{align}
Additionally, they integrated this approach into a trust region optimization framework to facilitate sampling from the state distribution $d^{\pi}$. 
\section{Constrained Model Agnostic Meta Reinforcement Learning}
\label{sec:Approach}
\begin{figure}
    \centering
    \includegraphics[width=\textwidth ,trim={0 0 0 6cm},clip]{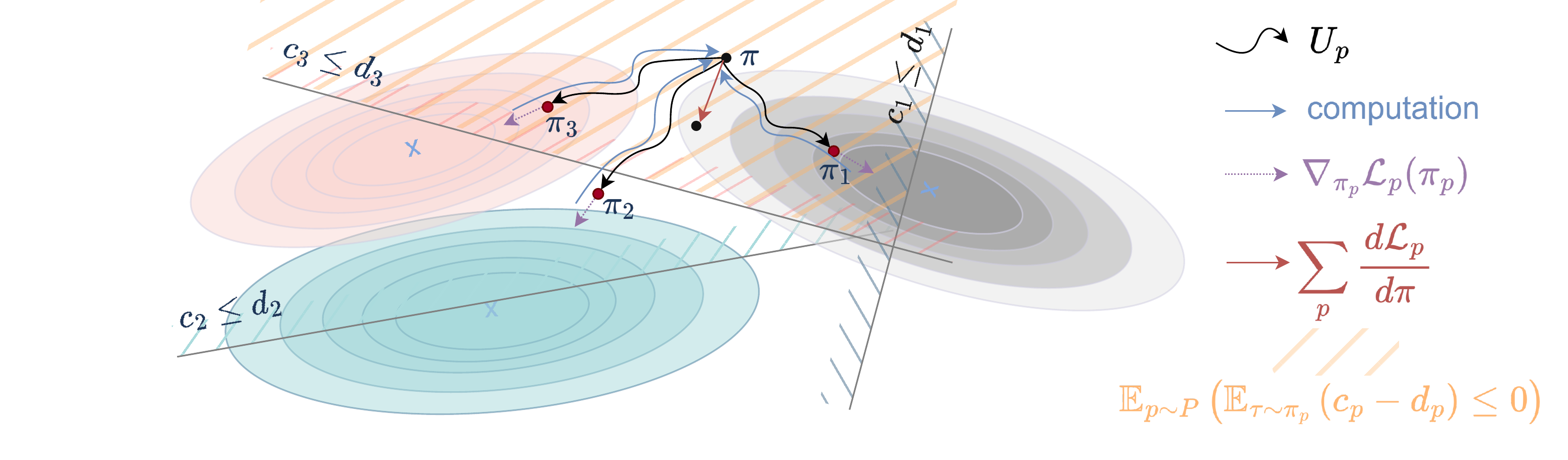}
    \caption{Visual representation of the Constrained Model Agnostic Meta Learning (C-MAML) framework. This schematic showcases the iterative optimization process where the meta-policy is trained across different tasks. Task-specific policies (\( \pi_1, \pi_2, \pi_3 \)) are adjusted within their respective constraint surfaces \( C_1, C_2, C_3 \), each with a dedicated safety boundary \( d_1, d_2, d_3 \).}
    \label{fig:enter-label}
\end{figure}
Our framework, Constrained Model Agnostic Meta Learning (C-MAML), introduces a gradient-based meta-learning approach designed to address both adaptability and safety in task learning. C-MAML incorporates task-specific constraints in the inner loop alongside a universal safety constraint in the outer loop that integrates all individual task constraints across the task distribution. This method ensures that each task's learning process is sensitive to its specific context and demands, embedding safety at the core of the adaptation process. Additionally, C-MAML's practical algorithm leverages first-order meta-gradient techniques combined with a global safety critic in the outer loop, aiming to strike a balance between computational efficiency and the safety of the resulting meta-policy when deployed.

\subsection{Incorporating Constraints in the Inner Loop}
Our framework models each task as a constrained Markov decision process (CMDP), aiming to optimize policies for maximum reward under specific constraints. This optimization occurs in the algorithm's inner loop, tailored for task-specific adjustments. The task-specific policy, denoted as \(\pi_p\), is obtained by solving the following optimization problem:
\begin{equation} \label{eq:MetaObjectiveInner}
  \pi_p = \arg\max_{\tilde{\pi}} \, \E_{\tau \sim \tilde{\pi}} \left( R_p(\tau | s_0 = s) \right) \quad \text{s.t.} \quad \E_{\tau \sim \tilde{\pi}} \left( C_p(\tau | s_0 = s) \right) - d_p \leq 0  
\end{equation}
One effective method for solving this CMDP is through Trust Region Policy Optimization with Lagrangian methods (TRPOLag). TRPOLag extends the robust performance guarantees of TRPO by integrating a Lagrangian framework that specifically addresses constraints. This integration enables the reformulation of the objective in Equation \eqref{eq:MetaObjectiveInner}
to optimize task-specific goals within defined safety constraints. The revised formulation is presented as follows:
\begin{alignat}{2}\label{eq:LagObjectiveInner}
    \mathcal{L}_{\textnormal{inner}} =\mathcal{L}_{p} (\pi_p, \lambda_p)=    \quad &\E_{\tau\sim\pi}\left(\sum_{t=0}^{\infty}\gamma^t\cdot\frac{\textcolor{blue}{\pi_p}(a_t|s_t)}{\pi(a_t|s_t)} \cdot A_p^{\pi}(s_t, a_t)\right) \\
    &-\lambda_p\cdot \left(\E_{\tau\sim\pi}\left(\sum_{t=0}^{\infty}\gamma^t\cdot\frac{\textcolor{blue}{\pi_p}(a_t|s_t)}{\pi(a_t|s_t)} \cdot A_{C_{p}}^{\pi}(s_t, a_t)\right) + J_C(\pi) - d\right)\nonumber\\
    &\text{s.t.} \quad D_{KL}(\textcolor{blue}{\pi_p} \|  \pi) \leq \epsilon \nonumber
\end{alignat}
This formulation utilizes dual variables, \(\lambda_p\), for each task \(p\). The parameter \(\epsilon\) serves as a predefined threshold for the KL divergence. For a detailed derivation of this equation, please refer to \ref{app:cmdp_trpoLag}.

\subsection{Optimizing Meta-Parameter in the Outer Loop}
Contrary to conventional Meta-RL algorithms that prioritize learning efficiency, our approach, C-MAML, integrates an intrinsic safety mechanism within the learning framework. This is accomplished by identifying a meta-parameter in the solution space that adheres to all constraints. This ensures that the selected initialization not only accelerates learning for newly encountered tasks in the inner loop but also complies with all constraints faced during the meta-training phase. This synthesis is encapsulated in the meta-objective function:
\begin{equation} \label{eq:MetaObjectiveOuter}
\max_\pi \, \E_{p \sim T} \left( \E_{\tau \sim \pi_p} \left( R_p(\tau | s_0 = s) \right) \right) \quad \text{s.t.} \quad \E_{p \sim T} \left( \E_{\tau \sim \pi_p} \left( C_p(\tau | s_0 = s) \right) - d_p \right) \leq 0
\end{equation}

\subsection{Enhancing Safety of the Meta-Policy}
 Employing MAML to find a meta-initialization that maximizes the objective function \eqref{eq:MetaObjectiveOuter} in the outer loop, introduces significant computational challenges due to its reliance on second-order gradients. This complexity is magnified when integrating inner loop constraints, making the outer loop's meta-policy \(\pi\) optimization notably intensive.  A practical solution is to utilize First-Order Model-Agnostic Meta-Learning (FoMAML), which simplifies the gradient estimation process by avoiding second-order derivatives. This approach enables more manageable updates to $\pi$ based on outcomes from task-specific policies. However, it is crucial to incorporate safety information derived during the inner loop's computations. Without this integration, the meta-policy might not consistently maintain safety standards across different task-specific policies despite its flexibility.
\begin{align}
\label{eq:CombinedProbWithEta}
\max_\pi\quad \E_{p\sim T}\left(\E_{\tau\sim{\textcolor{blue}{\pi_p}}}\left(R_p(\tau)\right)\right) \quad \text{s.t.} \quad & \E_{p\sim T}\left(\E_{\tau\sim{\textcolor{blue}{\pi_p}}}\left(C_p(\tau|s_0=s)\right) - d_p\right) \leq 0 \\
&\E_{p\sim T}\left( \E_{\tau\sim\textcolor{red}{\pi}}\left(C_p(\tau|s_0=s)\right) - d_p \right) \leq 0. \nonumber
\end{align}
This secondary constraint ensures that \(\pi\) consistently exhibits safe behavior across all tasks, leading to a modified Lagrangian optimization strategy for the outer loop:
\setlength{\jot}{-2pt} 
\begin{align}\label{eq:LagrangianProbWithEta}
\mathcal{L}_{\textnormal{outer}} (\pi, \lambda, \eta) = \E_{p\sim T}\Bigg[\E_{\tau\sim\textcolor{blue}{\pi_p}}\left(R_p(\tau|s_0=s)\right) 
& - \lambda \left(\E_{\tau\sim\textcolor{blue}{\pi_p}}\left(C_p(\tau|s_0=s)\right) - d_p\right) \\
& - \eta \left(\E_{\tau\sim\textcolor{red}{\pi}}\left(C_p(\tau|s_0=s)\right) - d_p\right)\Bigg].\nonumber 
\end{align}
Incorporating \(\eta\) as a crucial safety regulator enables the meta-policy \(\pi\) to efficiently adapt to new tasks while steadfastly upholding safe practices across different task environments.  Section \ref{sec:eta_eval} provides empirical validation of these benefits, demonstrating the significant advantage of integrating \(\eta\) into the outer loop optimization. Ensuring safety across the diverse range of tasks requires finding an initial setting (meta-parameter) that satisfies the unique constraints of each task. However, conflicting constraints among tasks make it challenging to identify a meta-parameter that is universally safe across the parameter space. To overcome this challenge and enable the discovery of such a meta-parameter, we refine the objective function \eqref{eq:LagrangianProbWithEta}. Initially configured for task-specific constraints $c_p$ and their respective thresholds $d_p$, our approach shifts focus towards adopting constraints $c$ and thresholds $d$ that are independent of any specific task. This transition is based on a logical foundation, considering universally relevant safety concerns. For example, the widespread constraint in robotics to prevent self-damage, such as avoiding excessive joint twisting, is pertinent and necessary across a broad spectrum of robotic tasks.
\setlength{\jot}{-2pt} 
\begin{align}\label{eq:LagrangianProbWithEtaConst}
\mathcal{L}_{\textnormal{outer}} (\pi, \lambda, \eta) = \E_{p\sim T}\Bigg[\E_{\tau\sim\pi_p}\left(R_p(\tau|s_0=s)\right) 
& - \lambda \left(\E_{\tau\sim\pi_p}\left(\textcolor{blue}{C}
(\tau|s_0=s)\right) - d\right) \\
& - \eta \left(\E_{\tau\sim\pi}\left(\textcolor{blue}{C}
(\tau|s_0=s)\right) - d\right)\Bigg]\nonumber.
\end{align}
Leveraging Assumption  \eqref{eq:LagrangianProbWithEtaConst}  enables the development of a meta cost critic, which can be trained with on-policy data gathered by the meta-policy across a variety of encountered tasks.
\begin{equation}
   V_C^\pi(s) = \E_{\tau\sim{\pi}}\left(C(\tau|s_0=s)\right). 
\end{equation}
Integrating the meta cost critic, the updated objective function is introduced as follows:
\begin{align}\label{eq:LagrangianProbWithEtaConstV}
\mathcal{L}_{\textnormal{outer}} (\pi, \lambda, \eta) =  \E_{p\sim T}\Bigg[\E_{\tau\sim\pi_p}\left(R_p(\tau|s_0=s)\right) 
& - \lambda \left(\E_{\tau\sim\pi_p}\left(C(\tau|s_0=s)\right) - d\right) 
 - \eta \left(\textcolor{blue}{V_C^\pi(s)} - d \right)\Bigg] .
\end{align}
Calculating the gradient of the objective function \eqref{eq:LagrangianProbWithEtaConstV} with respect to the meta-policy \(\pi\) returns:
\begin{align}
\nabla_\pi \mathcal{L}_{\textnormal{outer}}
&\approx  I\cdot \E_{p\sim T}\left(\nabla_\pi \mathcal{L}_{\textnormal{outer}}(\pi_p)\right) -  \frac{\partial}{\partial\pi} \eta \underbrace{\E_{\tau\sim \pi} \left( \sum_{t=0}^{\infty} \log(\pi(a_t|s_t)) \cdot V_c^\pi(s)\right)}_{\text{Independent of $\pi_p$}}
\end{align}
Here, the identity matrix $I$ is utilized as an approximation for the Hessian, simplifying the gradient computation process. The gradient  $\nabla_\pi \mathcal{L}_{\textnormal{outer}}(\pi_p)$ corresponds to FoMAML's return gradient, which balances rewards and costs. The second term, independent of task-specific updates \(\pi_p\), significantly affects the meta-policy's adaptation by promoting safer actions as evaluated by the safety critic. This effectively increases the propensity for actions to reduce costs and diminishes those associated with higher costs. A detailed gradient derivation is provided in Appendix \ref{app:practical_algo}, with a complete method diagram and pseudocode in Appendix \ref{app:practical_algo_diagram}.


\vspace{-2mm}
\section{Evaluation}
\label{sec:Evaluation}
The efficacy of Composite Model-Agnostic Meta-Learning (C-MAML) was evaluated through simulations involving high-dimensional mobile tasks with continuous state and action spaces. This analysis aimed to identify the practical merits and limitations of the approach. Key questions addressed during this evaluation include:
\begin{itemize}
    \setlength\itemsep{0em} 
    \item How does C-MAML ensure safety during \textbf{training} using first-order meta-learning and safety critic, and how does this impact policy safety and adaptability during \textbf{ fine-tuning}?
    \item What effects do safety constraints integrated into meta-initialization parameters have on adaptation speed and cost control in C-MAML, compared to traditional methods?
    \item Is C-MAML \textbf{agnostic} to the specific safe RL methods utilized in the inner loop?
\end{itemize}
\begin{figure}
    \centering
    \begin{subfigure}[t]{0.3\textwidth}
        \centering
        \includegraphics[width=\textwidth]{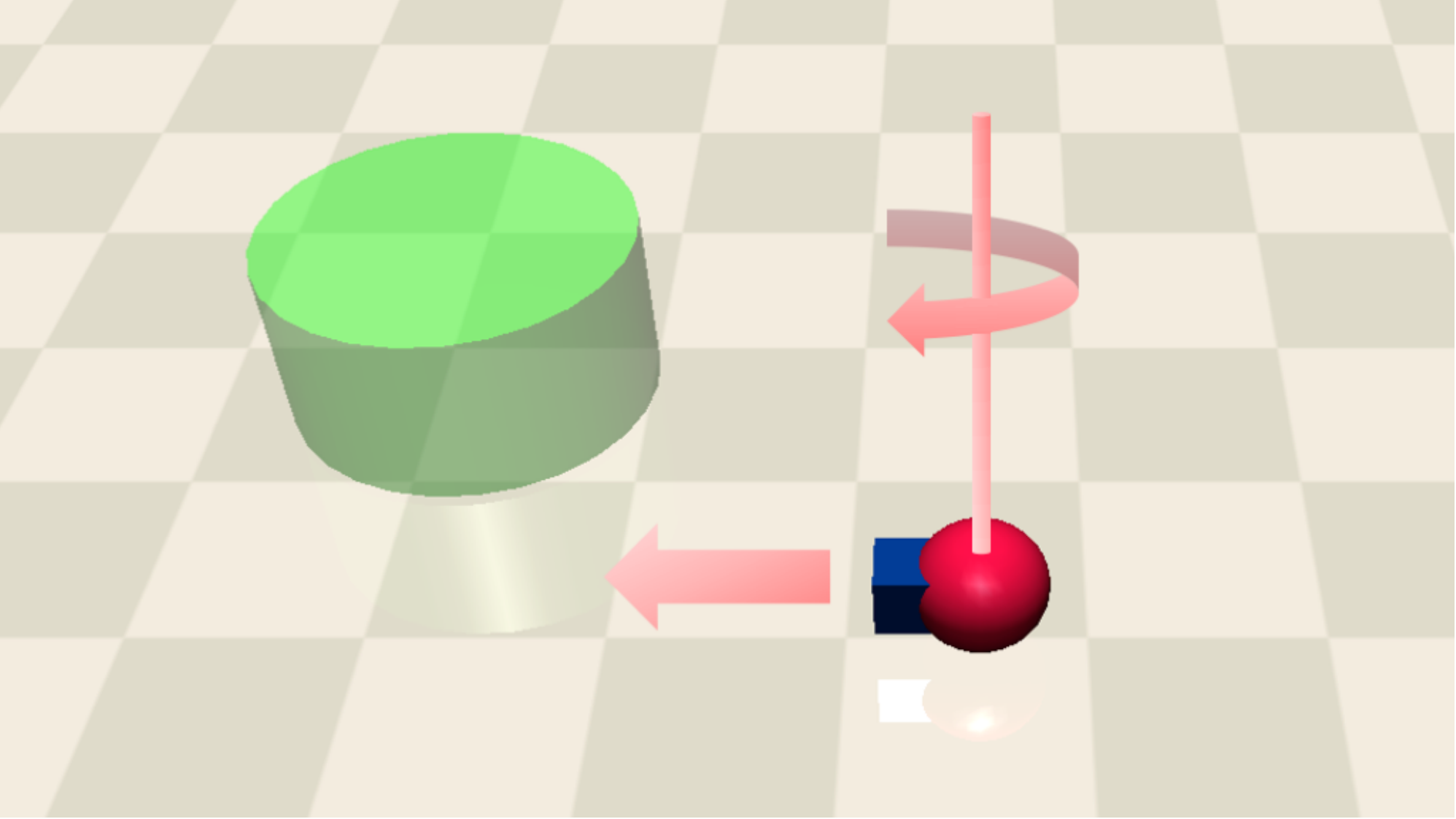}
        \caption{The point robot is shown without LiDAR markers. The arrows represent the two action dimensions. It has the task to reach the designated green region.}
        \label{fig:sub1}
    \end{subfigure}
    \hfill
    \begin{subfigure}[t]{0.3\textwidth}
        \centering
        \includegraphics[width=\textwidth]{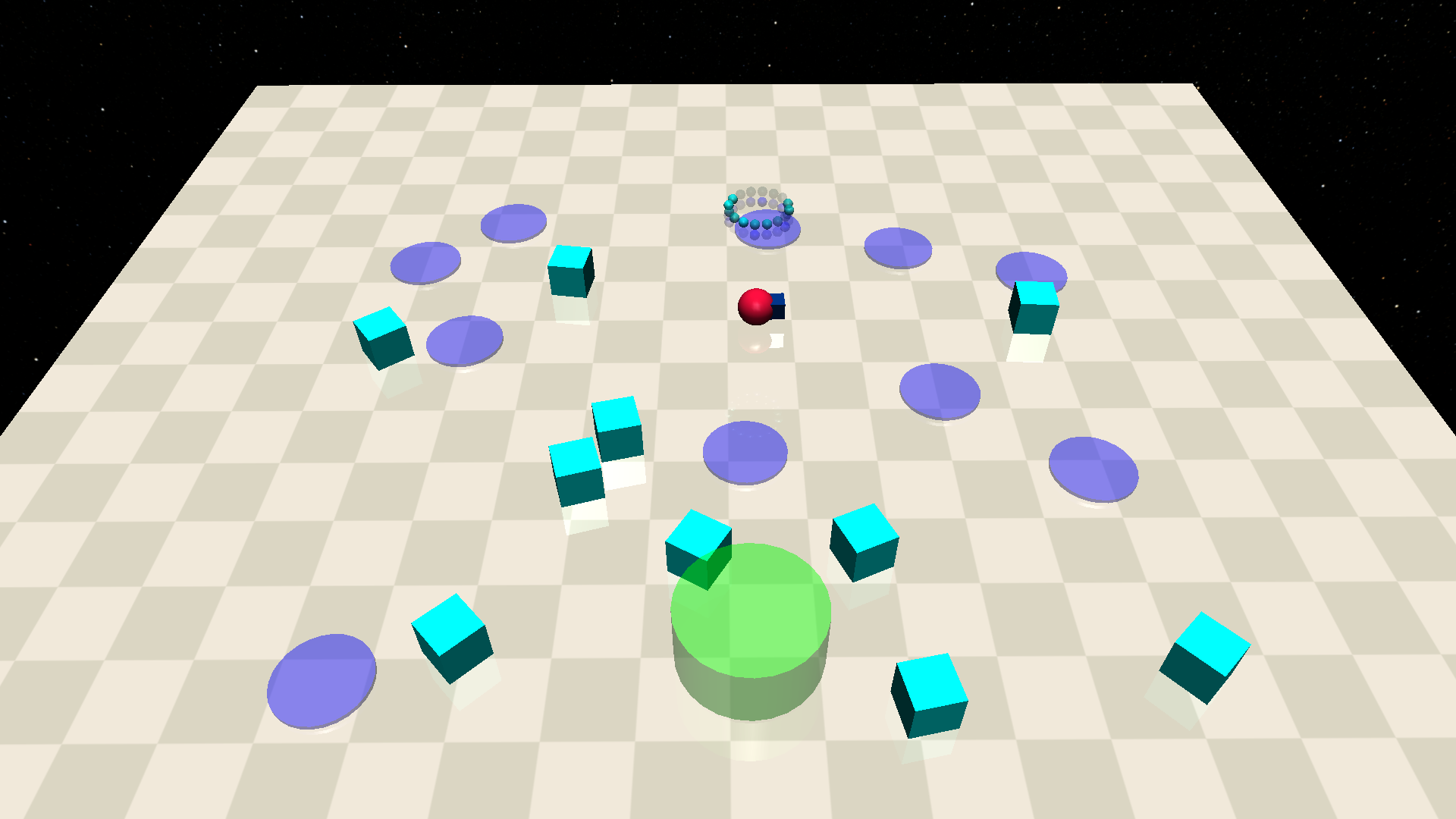}
        \caption{The environment is initialized with a specific task, controlling the configurations of obstacles and goals, as well as the agent's position.}
        \label{fig:sub2}
    \end{subfigure}
    \hfill
    \begin{subfigure}[t]{0.3\textwidth}
        \centering
        \includegraphics[width=\textwidth]{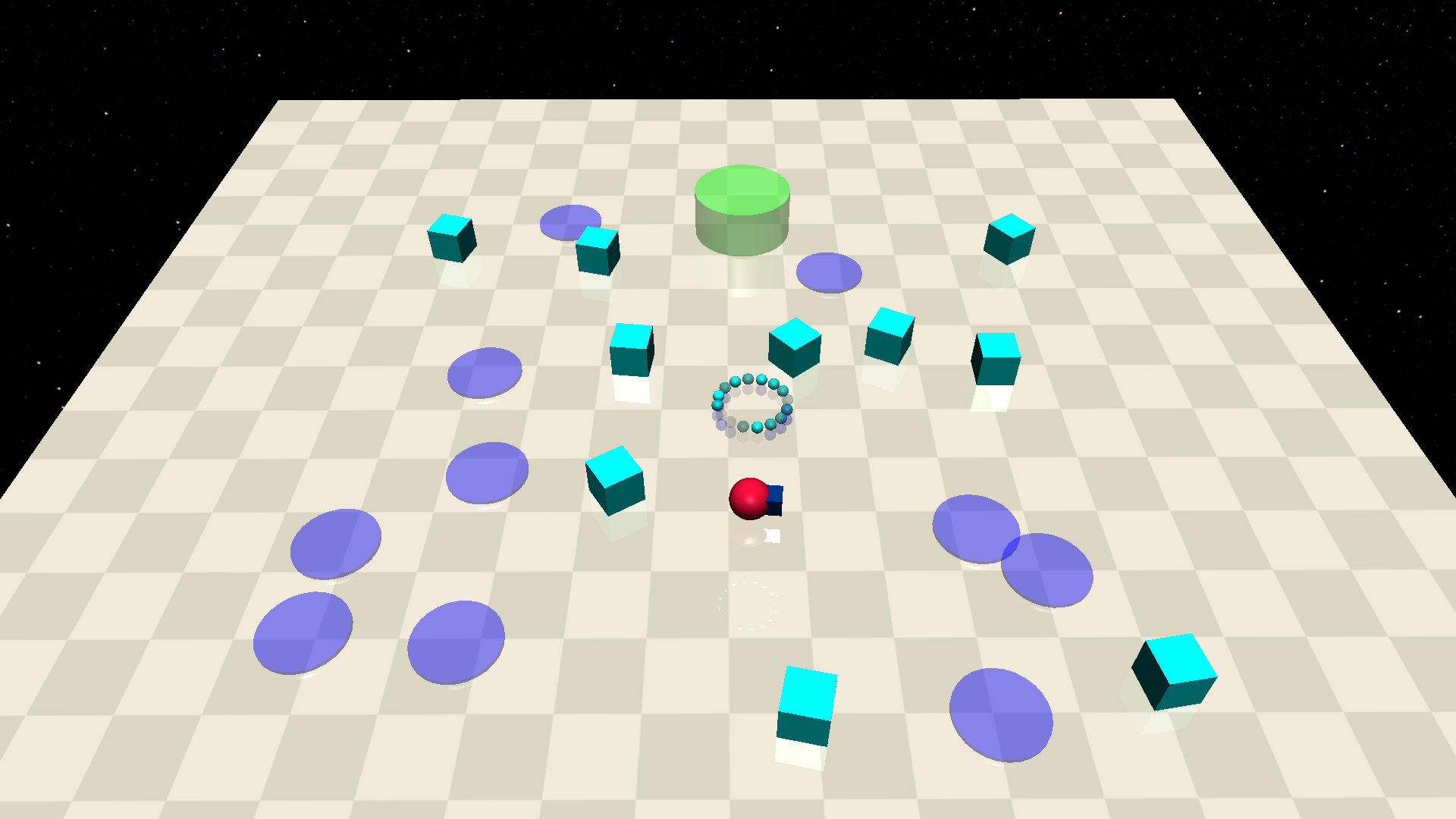}
        \caption{The variation of the task results in a different configuration of positions, yet the number of obstacles remains constant.}
        \label{fig:sub3}
    \end{subfigure}
    \caption{Illustrations of the action space and two different tasks of the used environment.}
    \label{fig:PointAgentSafetyGymnasium}
\end{figure}
\subsection{Task Setup}

\subsection{Importance of Using $\eta$ and Safety Critic} \label{sec:eta_eval}
\begin{figure}
    \centering
    \includegraphics[width=0.9\linewidth,trim={0 0.2cm 0 0},clip]{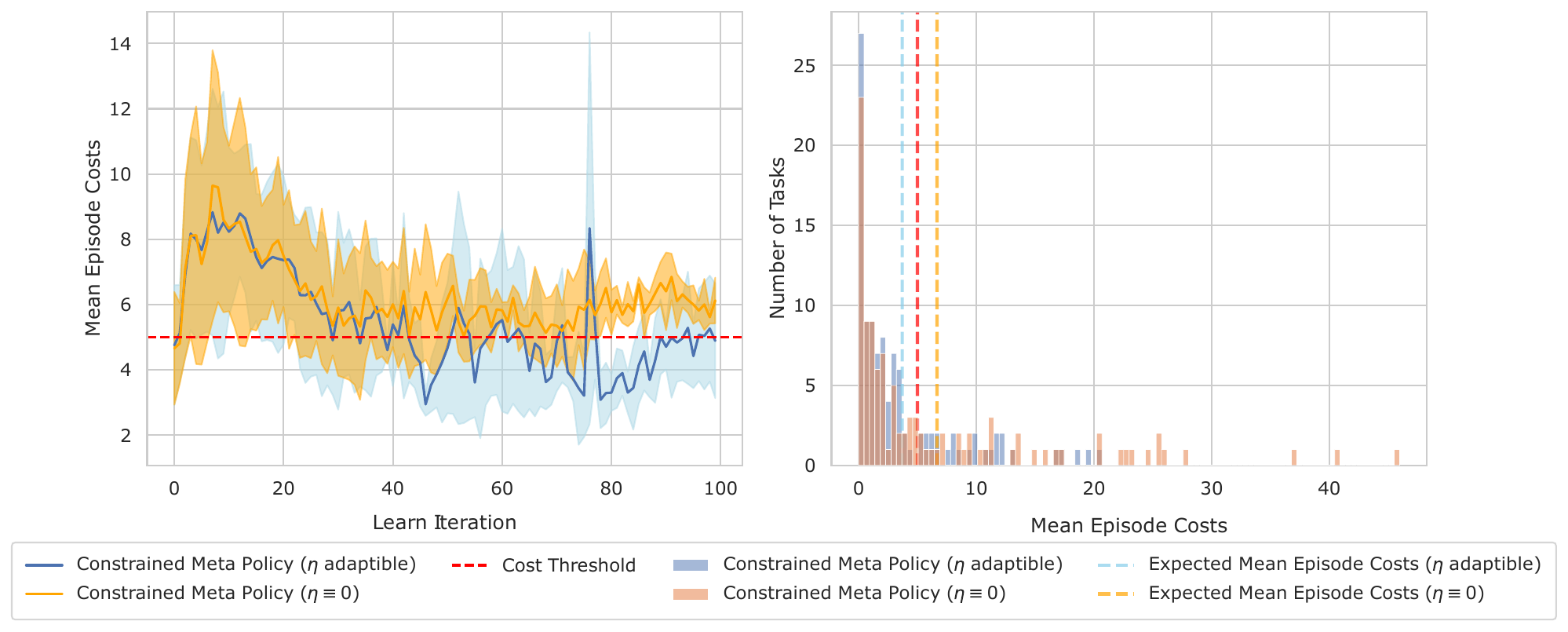}
    \caption{Evaluation of \(\eta\) on policy safety and adaptability: On the left, meta-training performance across 106 tasks, showing the effect of an adaptive \(\eta\) (employing safety critic) versus \(\eta = 0\) (no safety critic) on maintaining safer cost margins. On the right, fine-tuning phase performance, illustrating how an adaptive \(\eta\) contributes to consistent adherence to the \(d=5\) cost threshold compared to the absence of a safety critic.}
    \label{fig:eta-effect}
\end{figure}
\vspace{-3mm}
Utilizing First-Order Model Agnostic Meta Learning (FoMAML) instead of conventional MAML leads to the omission of valuable second-order information during the inner loop's optimization phase. To mitigate this, we introduce a new constraint, regulated by the Lagrange multiplier \(\eta\), into the optimization objective \eqref{eq:LagrangianProbWithEtaConstV}. Our evaluation centers on assessing the impact of \(\eta\) on policy safety and adaptability during both meta-training and fine-tuning stages. This involves a comparative analysis between two methodologies: one employing a trainable \(\eta\) and the other proceeding without it (Figure \ref{fig:eta-effect}). During meta-training, the adaptive \(\eta\), guided by the safety critic, helped maintain safer cost margins, avoiding the volatility observed with a fixed \(\eta\) at zero. In the fine-tuning phase, although both approaches occasionally breached the \(d=5\) cost limit, employing an adaptive \(\eta\) resulted in more consistent adherence to the cost threshold. This underscores the value of integrating an adaptive \(\eta\) and safety critic, ensuring safer policy adaptation and fine-tuning.

\subsection{Safety Adaptation Across Task Spectrum}
\begin{figure}
\centering
\includegraphics[width=0.7\linewidth,trim={0 0.22cm 0 0},clip]{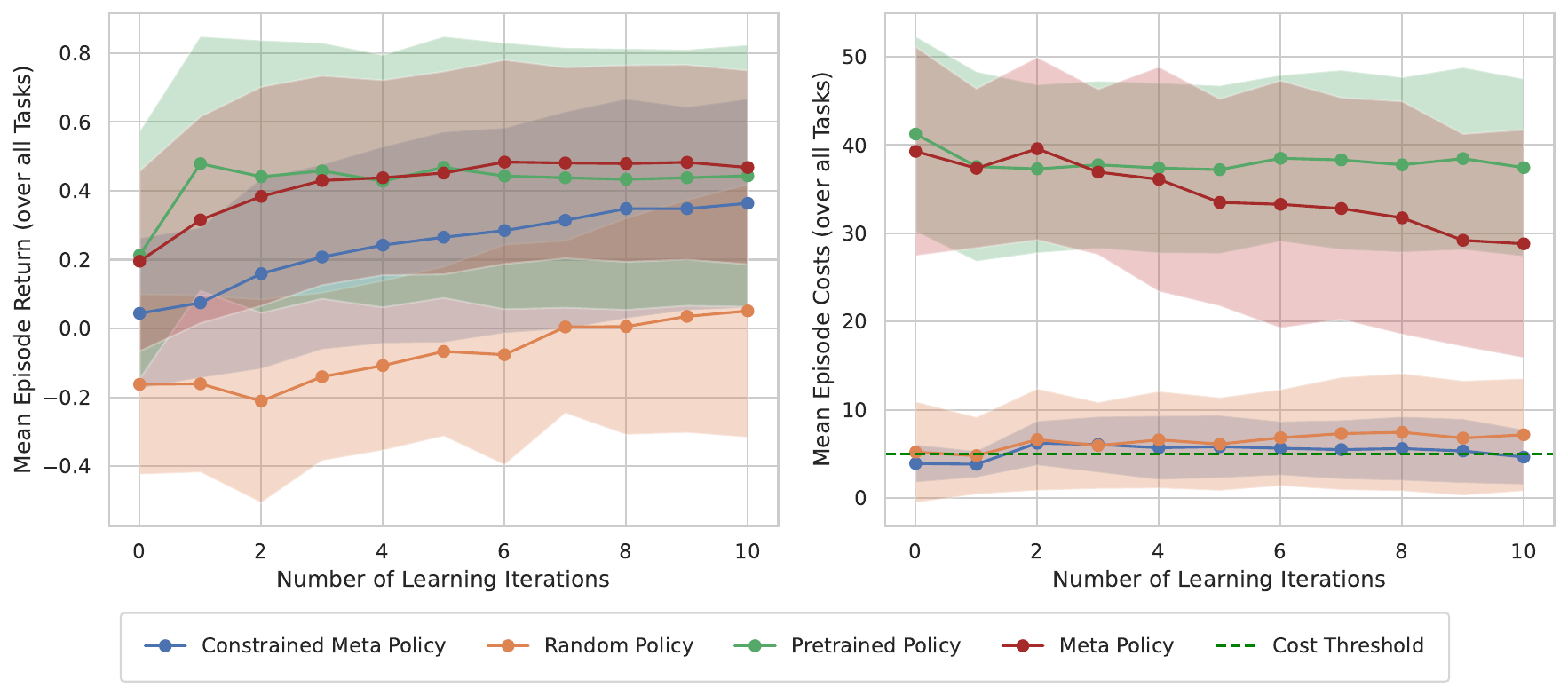}
\caption{Mean episode return and costs during fine-tuning across tasks. Policies are color-coded as follows: C-MAML with TRPOLag in the inner loop is depicted in blue, the randomly initialized policy in orange, the TRPOLag pretrained policy in green, and the MAML policy with TRPO in the inner loop is shown in red, highlighting the diverse adaptation strategies explored.}
\label{fig:TRPOLag_d5}
\end{figure}
\vspace{-3mm}
In the fine-tuning phase, the Trust Region Policy Optimization-Lagrangian (TRPOLag) algorithm was utilized, focusing on four policy initializations: randomly initialized, TRPOLag pretrained policy, MAML with TRPO in the inner loop, and C-MAML with TRPOLag in the inner loop. 
Our analysis across 106 tasks in Environment 2, shown in Figure \ref{fig:TRPOLag_d5}, includes episode returns and costs with confidence bands indicating standard deviation. Compared to random initialization, the C-MAML initialization achieved faster adaptation, maintaining episode costs close to the cost limit ($d=5$). Conversely, the TRPOLag pretrained policy and the unconstrained MAML initialization adapted more quickly but incurred higher costs during fine-tuning, highlighting their lack of safety considerations. This emphasizes the significance of incorporating safety into the adaptation process, as exemplified by C-MAML, to ensure both efficient and responsible policy development. Additional TRPOLag findings are detailed in Appendices \ref{App:CMAMLTRPOLag}, \ref{App:Rollouts} and \ref{App:DifficultyLevel}.
\vspace{-1mm}
\subsection{Model Agnosticity in C-MAML Framework}
\begin{figure}[H]
    \centering
    \includegraphics[width=0.7\linewidth,trim={0 0.2cm 0 0},clip]{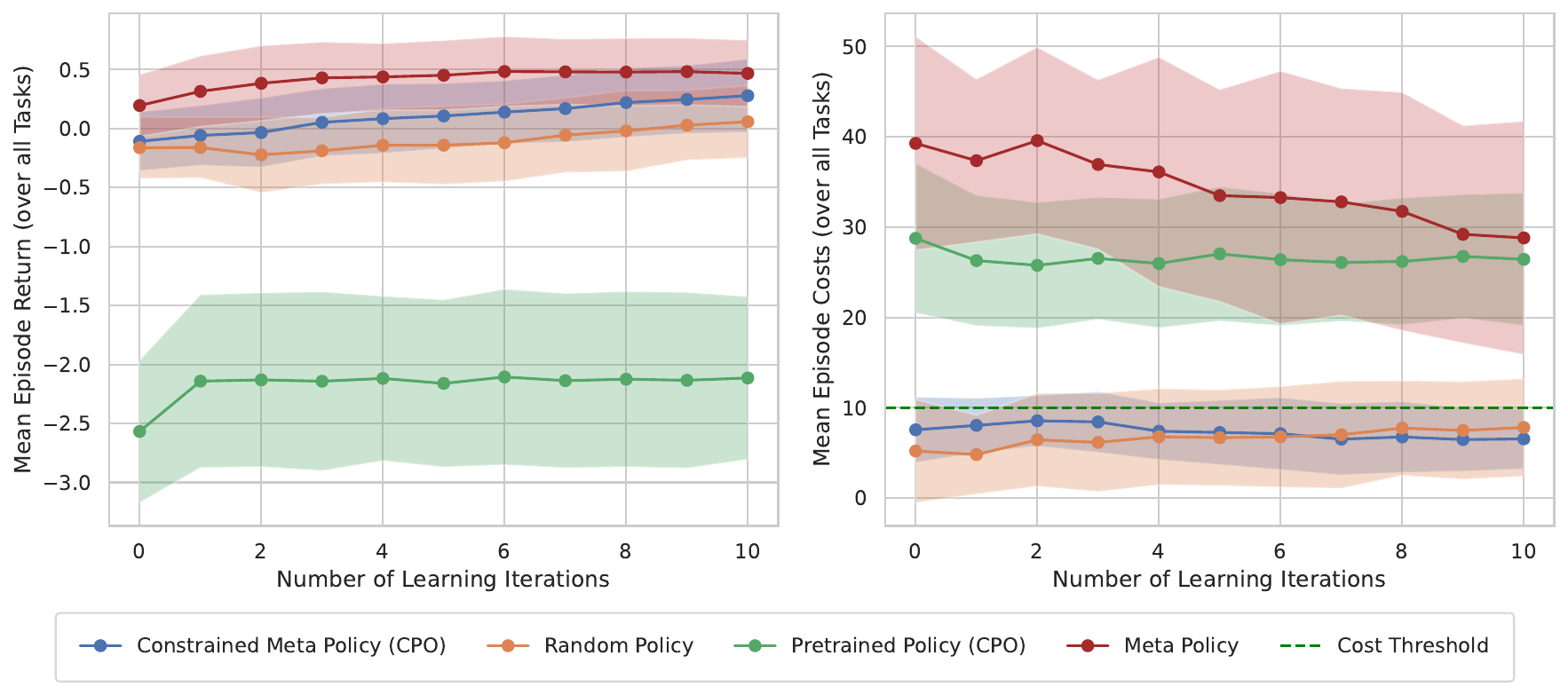}
    \caption{Mean episode return and costs during fine-tuning across tasks: C-MAML with CPO in the inner loop is depicted in blue, the randomly initialized policy in orange, CPO pretrained in green, and the MAML policy with TRPO in the inner loop is shown in red. Each of these policies (initialization) is fine-tuned using CPO.}    
    \label{fig:CPO_d10}
\end{figure}
\vspace{-3mm}
Adhering to the MAML concept of model agnosticism, our C-MAML framework is crafted to function independently from any particular algorithm within its inner optimization loop, facilitating the use of various constrained policy optimization strategies. To illustrate, we implemented Constrained Policy Optimization (CPO) as the inner loop algorithm. Results depicted in Figure \ref{fig:CPO_d10} reveal a performance trajectory consistent with those seen using C-MAML TRPOLag (Figure \ref{fig:TRPOLag_d5}), underscoring our framework's versatility and efficacy with different constrained policy optimization approaches. C-MAML CPO adheres to the \(d = 10\) cost limit, demonstrating efficient adaptability across tasks. For expanded insights on employing CPO in the inner loop, see Appendix \ref{App:CMAMLCPO}.

\vspace{-2mm}
\section{Conclusion}
\label{sec:Conclusion}
We presented Constrained Model-Agnostic Meta-Learning (C-MAML), a novel meta-RL framework that integrates gradient-based meta-learning with safe reinforcement learning to prioritize policy safety. Unlike conventional meta-RL algorithms that aim for efficient learning, C-MAML focuses on safe learning. It incorporates safe RL within its inner loop, resulting in policies that are adaptable and comply with safety norms. Our method enhances meta-policy safety by introducing additional constraints in the outer loop and employing a safety critic, ensuring adaptations prioritize safety. Our experiments demonstrate that C-MAML leads to safer and more adaptable policies compared to random or pre-trained starts. Future research could investigate other first-order meta-learning techniques and safe RL algorithms in the inner loop. 


\clearpage


\bibliography{example}  
\newpage
\begin{appendix}

\section{Notation}

To ensure clarity and precision in the equations throughout this document, we adopt the following notational conventions:

\begin{itemize}
    \item \( \tau \sim \pi \) indicates that the trajectory \( \tau \) is sampled according to the policy \( \pi \), where actions \( a \) are sampled from \( \pi \) and states \( s \) follow the discounted state distribution \( d^\pi \).
    \item When actions \( a \) and states \( s \) are sampled from different policies or distributions, we explicitly specify each source in the equations. For example, \( a \sim \pi \) and \( s \sim d^{\tilde{\pi}} \) indicates that actions are sampled according to policy \( \pi \) while states follow the distribution associated with a different policy \( \tilde{\pi} \).
\end{itemize}

\section{Mathematical Formulation of Constrained Meta Reinforcement Learning Objective}

In the domain of constrained Meta-RL, the overarching objective is to identify a meta-policy that is not only swiftly adaptable to new tasks but also maintains safety. The formal mathematical representation of this objective, along with the associated constraints, is detailed below.

\textbf{Primary Objective and Constraints:}
The primary goal in our approach is to maximize the expected cumulative discounted reward while ensuring that the accumulated discounted costs across tasks do not exceed a specific threshold. This is formalized in the following mathematical structure:
\begin{align}
\max_\pi\quad &\E_{p\sim T}\left(\E_{\substack{s_0\sim \mu_{p}\\a_t\sim\pi_p(\cdot|s_t)\\s_{t+1}\sim M(\cdot|s_t,a_t)}}\left(\sum_{t = 0}^{\infty} \gamma^t\cdot r_p(s_t,a_t,s_{t+1})\right)\right)\\
&s.t.\quad \E_{p\sim T}\left(\E_{\substack{s_0\sim \mu_{p}\\a_t\sim\pi_p(\cdot|s_t)\\s_{t+1}\sim M(\cdot|s_t,a_t)}}\left(\sum_{t = 0}^{\infty} \gamma^t\cdot c_p(s_t,a_t,s_{t+1})\right) - d_p\right) \leq 0.\nonumber
\end{align}
For clarity, \(R_p(\tau)\) represents the expected value of discounted trajectory returns for task \(p\), and \(C_p(\tau)\) indicates the expected value of discounted trajectory costs for the same task. The trajectories \(\tau\) are sampled from the policy \(\pi_p\), which is a task-specific policy derived from the general meta-policy \(\pi\).

The reformulated objective is then expressed as:
\begin{align} \label{equ:ObjeCMAML}
    \max_\pi\quad  &\E_{p\sim T}(\E_{\tau\sim \pi_p}(R_p(\tau)))\\
    s.t. \quad & \E_{p\sim T}(\E_{\tau\sim \pi_p}(C_p(\tau))) \leq d_p \nonumber
\end{align}
In this work, we consider a special case of the optimization problem mentioned above. Although problem \ref{equ:ObjeCMAML} accommodates task-specific cost functions \(c_p\) and thresholds \(d_p\), in practice, constraints that are uniform across all tasks are often more applicable. For instance, in robotic applications, there may be universal constraints such as preventing self-damage due to excessive joint twisting. Therefore, in the subsequent analysis, we explore a scenario with task-independent constraints, simplifying the model to accommodate a single, overarching constraint that applies uniformly to all tasks.
\begin{align} \label{equ:ObjeCMAML2}
    \max_\pi\quad  &\E_{p\sim T}(\E_{\tau\sim \pi_p}(R_p(\tau)))\\
    s.t.  \quad & \E_{p\sim T}(\E_{\tau\sim \pi_p}(C_p(\tau))) \leq d \nonumber
\end{align}

\textbf{Lagrangian Formulation in the Inner Loop:}
In our approach, the Lagrangian method plays a pivotal role in the inner loop optimization. Theoretically, if we consider \(\theta_p'\) as the actual stationary point, we encounter the following inner loop optimization problem:
\begin{align}
	\pi_p = \arg\min_{\lambda\geq 0}\max_{\tilde{\pi}}\quad \E_{\tau\sim \tilde{\pi}}(R_p(\tau)) - \lambda\E_{\tau\sim \tilde{\pi}}(C_p(\tau) - d)
\end{align}
Since \(\tilde{\pi}\) is an unknown policy to be determined, we can utilize the inequality by Kakade and Langford (\citeyear{KakadeLangford02}) to consider a surrogate function, akin to the one used in TRPO \citep{Schulman15TRPO}, which approximately corresponds to the above objective function. This eliminates the need to generate rollouts from a policy \(\tilde{\pi}\) to estimate the expected value.
\begin{alignat}{2}\label{equ:cMAML_UpdateInnerLoop}
\pi_p &= \arg \min_{\lambda\geq 0}\max_{\tilde{\pi}} \quad && \E_{\substack{s_t \sim d^{\tilde{\pi}} \\ a_t\sim\tilde{\pi}(\cdot|s_t)}} \left[ \sum_{t = 0}^{\infty} \gamma^t \cdot A^{\pi}(s_t, a_t) \right]\nonumber \\
& \quad && - \lambda \left( \E_{\substack{s_t \sim d^{\tilde{\pi}} \\ a_t\sim\tilde{\pi}(\cdot|s_t)}} \left[ \sum_{t = 0}^{\infty} \gamma^t \cdot A_C^{\pi}(s_t, a_t) \right] - d \right) \nonumber\\
&\stackrel{\substack{\text{Importance}\\\text{Sampling}}}{=} \arg \min_{\lambda\geq 0}\max_{\tilde{\pi}} \quad && \E_{\substack{s_t \sim d^{\tilde{\pi}} \\ a_t\sim\pi(\cdot|s_t)}} \left( \sum_{t = 0}^{\infty} \gamma^t \cdot \frac{\tilde{\pi}(a_t|s_t)}{\pi(a_t|s_t)} \cdot A^{\pi}(s_t, a_t) \right) \nonumber \\
& \quad && - \lambda  \left( \E_{\substack{s_t \sim d^{\tilde{\pi}} \\ a_t\sim\pi(\cdot|s_t)}} \left( \sum_{t = 0}^{\infty} \gamma^t \cdot \frac{\tilde{\pi}(a_t|s_t)}{\pi(a_t|s_t)} \cdot A_C^{\pi}(s_t, a_t) \right) - d \right) \nonumber\\
&\stackrel{\textnormal{TRPO}}{\approx} \arg \min_{\lambda\geq 0}\max_{\pi}\quad && \E_{\substack{s_t \sim d^{\pi} \\ a_t\sim\pi(\cdot|s_t)}}\left(\sum_{t = 0}^{\infty} \gamma^t\cdot \frac{\tilde{\pi}(a_t|s_t)}{\pi(a_t|s_t)}\cdot A^{\pi}(s_t,a_t)\right)\\
& \quad && - \lambda \left( \E_{\substack{s_t \sim d^{\pi} \\ a_t\sim\pi(\cdot|s_t)}} \left(\sum_{t = 0}^{\infty} \gamma^t\cdot \frac{\tilde{\pi}(a_t|s_t)}{\pi(a_t|s_t)}\cdot A_C^\pi(s_t,a_t)\right) - d\right)\nonumber\\
& s.t.  \quad D_{KL}(\tilde{\pi},\pi)\leq\epsilon\nonumber
\end{alignat}
By employing a primal-dual gradient approach, the optimization can be iteratively performed for both \(\tilde{\pi}\) and \(\lambda\). This optimization problem must be resolved in the inner loop of the MAML-TRPOLag algorithm for each selected task \(p\). 

\textbf{First Order Optimization in Outer Loop:}
Transforming equation \eqref{equ:ObjeCMAML2} from constrained to unconstrained optimization using the Lagrangian method, we derive the following outer loop optimization problem:
\begin{align} \label{equ:}
\mathcal{L}_{\textnormal{outer}} = \E_{p\sim T}\left(\E_{\tau\sim \pi_p}(R_p(\tau)) -\lambda\left( \E_{\tau\sim \pi_p}\left(C_p(\tau) - d \right)\right) \right)
\end{align}
This equation reveals the presence of two meta-parameters: the policy parameter $\pi$ and the Lagrange multiplier $\lambda$. Therefore, it is necessary to compute the partial derivative for each of these parameters.

The calculation of the meta-gradient with respect to the meta-policy $\pi$ is as follows:
\begin{align*}
	\nabla_\pi \mathcal{L}_{\textnormal{outer}}
	&= \frac{\partial}{\partial\pi} \E_{p\sim T}\left(\E_{\tau\sim \pi_p}(R_p(\tau)) -\lambda\left( \E_{\tau\sim \pi_p}\left(C_p(\tau) - d \right)\right)  \right).\\
	&= \frac{\partial\pi_p}{\partial\pi}\frac{\partial}{\partial\pi_p}  \E_{p\sim T}\left(\E_{\tau\sim \pi_p}(R_p(\tau)) -\lambda\left( \E_{\tau\sim \pi_p}\left(C_p(\tau) - d \right)\right)  \right)\\
	&= \frac{\partial U_p(\pi)}{\partial\pi}\cdot \E_{p\sim P}\left(\frac{\partial}{\partial\pi_p} \E_{\tau\sim \pi_p}(R_p(\tau)) - \lambda\cdot\frac{\partial}{\partial\pi_p} \E_{\tau\sim \pi_p}\left(C_p(\tau) - d \right)\right) \\
	&\stackrel{\substack{\textnormal{Policy}\\\textnormal{Gradient}}}{=}\frac{\partial U_p(\pi)}{\partial\pi}\cdot \E_{p\sim T}\left(\E_{\tau\sim \pi_p}\left(\sum_{t = 0}^{\infty} \nabla_{\pi_p}\log(\pi_p(a_t|s_t)) \cdot R_p(\tau)\right)
	- \lambda\cdot \E_{\tau\sim \pi_p}\left(\sum_{t = 0}^{\infty} \nabla_{\pi_p}\log(\pi_p(a_t|s_t)) \cdot C_p(\tau)\right)\right)\\
\end{align*}
In the First order MAML (FoMAML) framework, the derivative of the update rule \( U_p(\pi) \) is approximated as an identity matrix, simplifying the gradient computation. This approximation is essential for efficiently updating the meta-policy, aligning with the objective and constraints.
\begin{align} \label{equ:delta_louter_pi}
\nabla_\pi \mathcal{L}_{\textnormal{outer}}
	&= \textcolor{blue}{\frac{\partial U_p(\pi)}{\partial\pi}}\cdot \E_{p\sim T}\left(\E_{\tau\sim \pi_p}\left(\sum_{t = 0}^{\infty} \nabla_{\pi_p}\log(\pi_p(a_t|s_t)) \cdot R_p(\tau)\right)
	- \lambda\cdot \E_{\tau\sim \pi_p}\left(\sum_{t = 0}^{\infty} \nabla_{\pi_p}\log(\pi_p(a_t|s_t)) \cdot C_p(\tau)\right)\right)\nonumber\\
    &\stackrel{\textnormal{FoMAML}}{\approx} \textcolor{blue}{I}\cdot \E_{p\sim T}\left(\E_{\tau\sim \pi_p}\left(\sum_{t = 0}^{\infty} \nabla_{\pi_p}\log(\pi_p(a_t|s_t)) \cdot R_p(\tau)\right)
	- \lambda\cdot \E_{\tau\sim \pi_p}\left(\sum_{t = 0}^{\infty} \nabla_{\pi_p}\log(\pi_p(a_t|s_t)) \cdot C_p(\tau)\right)\right)
\end{align}
For the meta Lagrange parameter $\lambda$, the gradient is calculated as follows:
\begin{align} \label{equ:delta_louter_lambda}
\nabla_\lambda \mathcal{L}_{\textnormal{outer}}
&= \frac{\partial}{\partial\lambda} \E_{p\sim T}\left(\E_{\tau\sim \pi_p}(R_p(\tau)) -\lambda\left( \E_{\tau\sim \pi_p}\left(C_p(\tau) - d \right)\right)  \right)\nonumber\\
&= - \E_{p\sim T}\left(\E_{\tau\sim \pi_p}\left(C_p(\tau) - d \right)\right), 
\end{align}
such that the constraint $\lambda\geq 0$ is maintained.

\section{Solving CMDP with TRPO and Lagrangian Methods}
\label{app:cmdp_trpoLag}

In the domain of Constrained Markov Decision Processes (CMDP), we aim to optimize policies that maximize expected rewards while adhering to predefined constraints on expected costs. The typical CMDP problem is formulated as follows:
\begin{equation}
\label{eq:ObjectMDP}
    \pi^* = \arg\max_{\tilde{\pi}} \E_{\tau \sim \tilde{\pi}}[R(\tau | s_0 = s)] \quad \text{s.t.} \quad \E_{\tau \sim \tilde{\pi}}[C(\tau | s_0 = s)] \leq d,
\end{equation}
where \( \pi^* \) denotes the optimal policy, \( R(\tau | s_0 = s) \) and \( C(\tau | s_0 = s) \) represent the total reward and cost from trajectory \( \tau \), respectively, and \( d \) is the allowable cost threshold.

To transform this CMDP into an unconstrained optimization problem, we employ the Lagrangian method by introducing a Lagrange multiplier \( \lambda \):
\begin{equation}
    \mathcal{L}(\pi, \lambda) = \E_{\tau \sim \pi}[R(\tau | s_0 = s)] - \lambda \left(\E_{\tau \sim \pi}[C(\tau | s_0 = s)] - d\right),
\end{equation}
where \( \lambda \) penalizes the policy when the expected cost exceeds the threshold \( d \).

\subsection{Trust Region Policy Optimization (TRPO)}

Trust Region Policy Optimization (TRPO) addresses the optimization of policies in Markov Decision Processes (MDPs) by introducing a KL divergence constraint. This constraint ensures that policy updates remain within a predefined "trust region", thereby promoting stability. The objective for unconstrained policy improvement using TRPO is given by:
\begin{equation}
   \pi^* = \arg\max_{\tilde{\pi}}  
   \E_{\tau \sim\pi}\left[\sum_{t=0}^{\infty}\gamma^t \cdot \frac{\textcolor{blue}{\tilde{\pi}}(a_t|s_t)}{\pi(a_t|s_t)} \cdot A^{\pi}(s, a)\right] \quad \text{s.t.} \quad D_{KL}(\pi \|\tilde{\pi}) \leq \epsilon,
\end{equation}
where \( A^{\pi}(s, a) \) is the advantage function of the old policy \( \pi \), and \( \delta \) denotes the size of the trust region as determined by the KL divergence threshold.
\subsection{Incorporating Cost in Policy Optimization Strategies}

Building upon TRPO, \cite{Achiam17CPO} introduced an extended framework that accommodates policy constraints. This approach equates the policy's performance regarding rewards to a similar measure for the expected value of accumulated future costs:
\begin{align*}
    J_C(\pi) = \E_{\tau\sim \pi}\left(C(\tau) - d \right).
\end{align*}
The performance concerning costs can also be approximated as:
\begin{align*}
     J_C(\tilde{\pi}) = J_C(\pi) + \E_{\substack{a\sim\pi\\s\sim d^{\textcolor{blue}{\tilde{\pi}}}}}\left(\sum_{t=0}^{\infty}\gamma^t\cdot\frac{\textcolor{blue}{\tilde{\pi}}(a_t|s_t)}{\pi(a_t|s_t)} \cdot A_C^{\pi}(s_t, a_t)\right),
\end{align*}

where \( A^{\pi}_c(s, a) \) is the advantage cost function of the old policy \( \pi \).

Assuming the same approximation of discounted state distribution $d^{\tilde{\pi}}$ as $d^{\pi}$ given the KL divergence constraint between policies, the resulting optimization problem becomes:

\begin{alignat}{2}\label{equ:CPOOptprob}
    \max_{\tilde{\pi}} \quad &E_{\substack{a\sim\pi\\s\sim d^{\pi}}}\left(\sum_{t=0}^{\infty}\gamma^t\cdot\frac{\textcolor{blue}{\tilde{\pi}}(a_t|s_t)}{\pi(a_t|s_t)} \cdot A^{\pi}(s_t, a_t)\right)\\
    \text{s.t.} \quad &J_C(\pi) + E_{\substack{a\sim\pi\\s\sim d^{\pi}}}\left(\sum_{t=0}^{\infty}\gamma^t\cdot\frac{\textcolor{blue}{\tilde{\pi}}(a_t|s_t)}{\pi(a_t|s_t)} \cdot A_C^{\pi}(s_t, a_t)\right) &&\leq d\nonumber\\
    & D_{KL}(\textcolor{blue}{\tilde{\pi}} \| \pi) \leq \delta.\nonumber
\end{alignat}

\subsection{Integrating TRPO with Lagrangian (TRPOLag)}

A further extension is discussed in Ray et al. (2019) \cite{SafetyGymOpenAI}, which employs the Lagrangian method to simplify CMDP solutions. This adaptation modifies the TRPO objective to incorporate both reward maximization and constraint adherence by adjusting the Lagrange multiplier $\lambda$:

\begin{alignat}{2}\label{eq:LagObjectiveInner}
    \mathcal{L} (\pi, \lambda)=    \quad &\E_{\tau\sim\pi}\left(\sum_{t=0}^{\infty}\gamma^t\cdot\frac{\textcolor{blue}{\tilde{\pi}}(a_t|s_t)}{\pi(a_t|s_t)} \cdot A_p^{\pi}(s_t, a_t)\right) \\
    &-\lambda_p\cdot \left(\E_{\tau\sim\pi}\left(\sum_{t=0}^{\infty}\gamma^t\cdot\frac{\textcolor{blue}{\tilde{\pi}}(a_t|s_t)}{\pi(a_t|s_t)} \cdot A_{c_{p}}^{\pi}(s_t, a_t)\right) + J_C(\pi) - d\right)\nonumber\\
    & \text{s.t.} \quad D_{KL}(\textcolor{blue}{\tilde{\pi}} \| \pi)\leq \epsilon\nonumber
\end{alignat}
where \( A^{\pi_{\text{old}}}_R(s, a) \) and \( A^{\pi_{\text{old}}}_C(s, a) \) are the advantage functions for the reward and cost, respectively.

This approach allows TRPOLag to effectively balance the trade-offs between maximizing rewards and adhering to cost constraints, providing a robust solution framework for CMDPs.

\section{Mathematical Formulation of the Practical Algorithm} 
\label{app:practical_algo}
Our practical algorithm adapts the objective function \eqref{equ:ObjeCMAML2} of the method by adding a new constraint to ensure that the new meta-policy adheres to universal constraints.
\begin{align} \label{equ:ObjeCMAML3}
    \max_\pi\quad  &\E_{p\sim T}(\E_{\tau\sim \pi_p}(R_p(\tau)))\\
    s.t.  \quad & \E_{p\sim T}(\E_{\tau\sim \pi_p}(C_p(\tau)) \leq d) \nonumber\\
    \quad & \E_{p\sim T}(\E_{\tau\sim \pi}(C_p(\tau)) \leq d) \nonumber
\end{align}

By rewriting Equation \eqref{equ:ObjeCMAML3} using Lagrangian formulation, we obtain an unconstrained objective. Additionally, by using the meta safety critic as an approximation for the expected cumulative sum of the meta-policy, we arrive at the following equation. 

\begin{align} \label{equ:ObjeCMAML4}
    \mathcal{L}_{\textnormal{outer}} = \E_{p\sim T}\left(\E_{\tau\sim \pi_p}(R_p(\tau))\right)
    -\lambda \E_{p\sim T}\left(\E_{\tau\sim \pi_p}(C_p(\tau))- d \right) -  \eta\E_{\tau\sim \pi} \left(V_c^\pi(s) \right) - d 
\end{align}

The gradient of the objective function with respect to the meta-policy \(\pi\) is approximated as follows:

\begin{align}
\nabla_\pi \mathcal{L}_{\textnormal{outer}}
&\approx I\cdot \E_{p\sim T}\left(\underbrace{\E_{\tau\sim \pi_p}\left(\sum_{t = 0}^{\infty} \nabla_{\pi_p}\log(\pi_p(a_t|s_t)) \cdot R_p(\tau)\right)
- \lambda\cdot \E_{\tau\sim \pi_p}\left(\sum_{t = 0}^{\infty} \nabla_{\pi_p}\log(\pi_p(a_t|s_t)) \cdot C_p(\tau)\right)}_{\text{FoMAML's return gradient balancing rewards and costs}}\right)\nonumber\\
 & -  \frac{\partial}{\partial\pi} \eta \underbrace{\E_{\tau\sim \pi} \left( \sum_{t=0}^{\infty} \log(\pi(a_t|s_t)) \cdot V_c^\pi(s)\right)}_{\text{Independent of $\pi_p$}}
\end{align}

In this formulation, the first term under the approximation symbol represents the gradient derived from FoMAML, which balances rewards and costs. The second term emphasizes the contribution of the meta safety critic \(V_c^\pi(s)\) to the overall gradient with respect to \(\pi\).

For the meta Lagrange parameter $\lambda$, the gradient is calculated as in the equation \eqref{equ:delta_louter_lambda}.   An additional gradient is computed to optimize $\eta$:
\begin{align} 
\nabla_\eta \mathcal{L}_{\textnormal{outer}} = 
 \E_{\tau\sim \pi} \left(V_c^\pi(s) \right) - d 
\end{align} 
\section{C-MAML with $\eta$ and Safety Critic}
\label{app:practical_algo_diagram}
\begin{figure}[H]
    \centering
    \includegraphics[width=0.5\linewidth, trim={0 0.7cm 0.3cm 0}, clip]{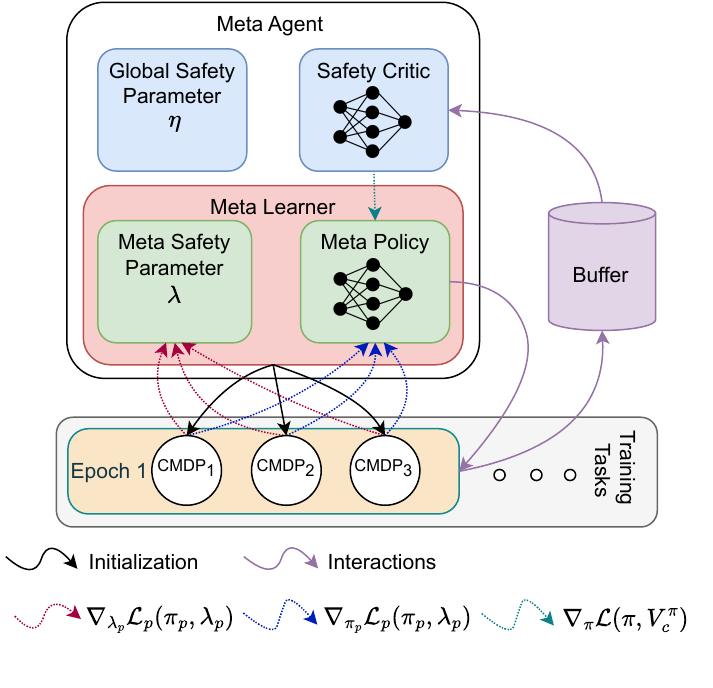}
    \caption{Diagram depicting the TRPOLag-based algorithm's execution: Initialization involves setting up task-specific policies \(\pi_p\) and safety parameters \(\lambda_p\) for each $\text{CMPD}_p$, derived from the meta-policy \(\pi\) and meta-parameter \(\lambda\) (\textbf{black arrows}). Data collection for each task employs these initial parameters (\textbf{purple arrows}). This data informs the refinement of the meta safety critic and the task-specific policy updates \(\pi_p\). Feedback from task performances, encapsulating both reward and safety, guides the meta-policy \(\pi\) updates (\textbf{blue dotted arrows}). Simultaneously, the meta-policy is reinforced by the meta safety critic to align with overarching constraints (\textbf{green dotted arrows}), while aggregate signals from \(\lambda_p\) adjustments inform the meta-parameter \(\lambda\) updates (\textbf{red dotted arrows}).}
    \label{fig:Cmaml}
\end{figure}

The practical algorithm utilizes a first-order meta-learning approach, integrating safety-critic with the Lagrangian multiplier $\eta$ for enhanced safety. This is illustrated in Figure \ref{fig:Cmaml}.
 The pseudocode detailing our method is presented below.
\begin{algorithm}
\caption{Constrained Model Agnostic Meta Learning (C-MAML)}
\label{alg:MetaConstrainedPolicyOpt}
\begin{algorithmic}[1]
\State \textbf{Input:} Number of outer iteration steps $N$, number of tasks for computing the meta-gradient $B$
\State \textbf{Output:} Meta-Policy $\pi_{\theta}$ and Cost Critic $V_{C,\hat{\theta}}$
\State Initialize the meta-policy $\pi_{\theta}$ and the meta-cost critic $V_{C,\hat{\theta}}$ randomly
\For{counter $\leq N$}
\State Randomly select a subset $\mathcal{P}$ of the support of $P$ with $|\mathcal{P}| = B$
\For{task $p$ in $\mathcal{P}$}
\State $\pi_p' := \pi_{\theta}, V_{C,p} := V_{C,\hat{\theta}}$
\State Randomly initialize the Reward Critic $V_{R,p}$
\For{number of adaptation steps}
\State Generate rollouts with $\pi_p'$
\State Update the policy $\pi_p'$ using a constrained optimization algorithm
\State Update the critics $V_{C,p}$ and $V_{R,p}$
\EndFor
\State Compute the meta-gradient for task $p$ for the meta-policy and the meta-cost critic
\EndFor
\State Average the task-specific meta-gradients and update $\pi_{\theta}$ and $V_{C,\hat{\theta}}$ using a meta-learning algorithm
\EndFor
\State \Return $\pi_{\theta}, V_{C,\hat{\theta}}$
\end{algorithmic}
\end{algorithm}
\section{Task Distributions} 
The positions of the agent, obstacles, and the goal vary randomly with each task. The seeds were chosen solely to require the agent to search for goals within a 140-degree arc.

\newpage
\subsection{Environment 1}\label{App:Env1}
\begin{figure}[H]
    \centering
    \includegraphics[width=\linewidth, trim=7cm 5cm 5cm 2cm, clip]{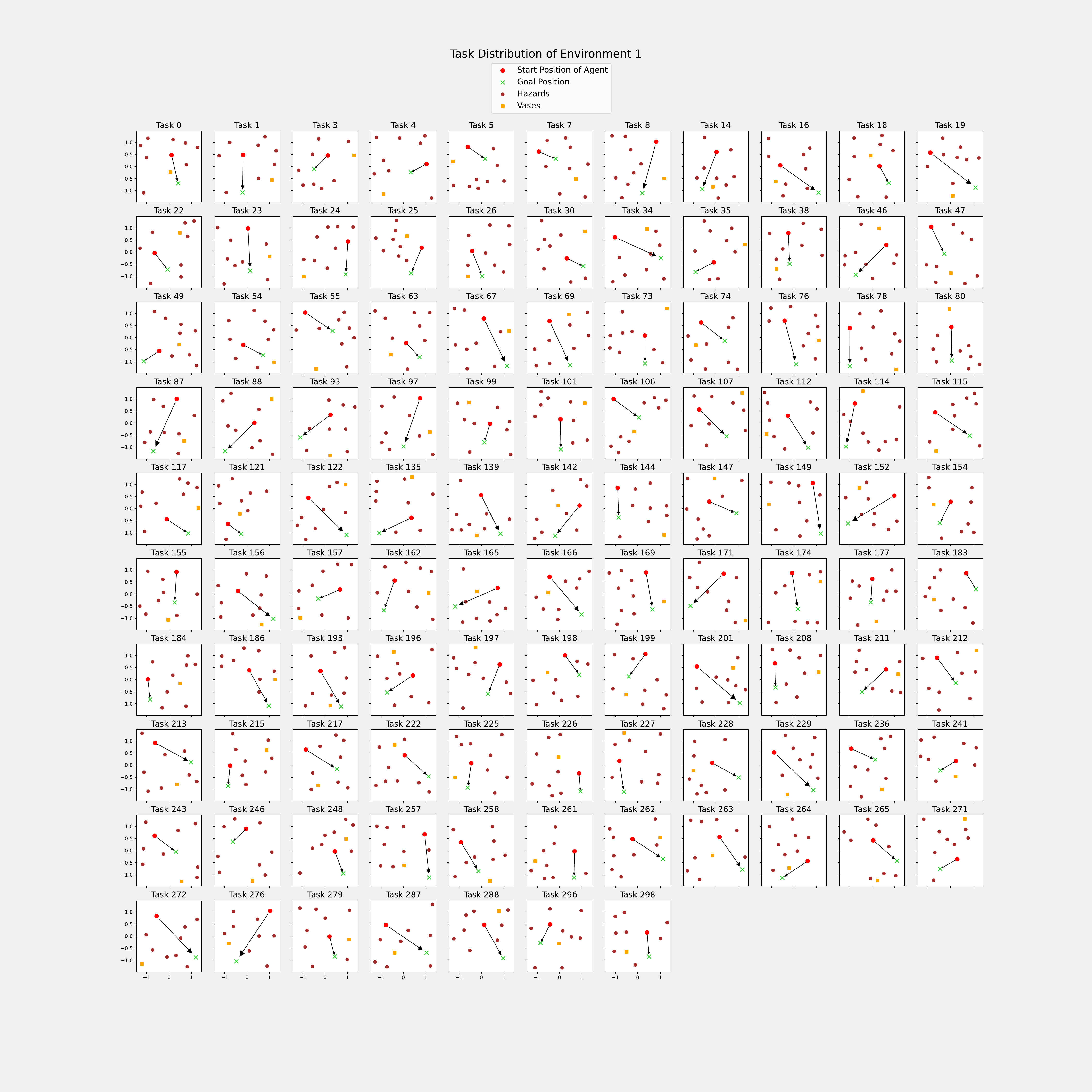}
    \caption{All 107 seeds of the task distribution under Environment 1 are depicted. The tasks were drawn uniformly without replacement.}
    \label{fig:TaskDistrEnv1}
\end{figure}

Environment 1 encompasses 107 tasks, illustrated in Figure \ref{fig:TaskDistrEnv1}, representing all seeds within this arc in the seed range of 0 to 300. The number of hazards is limited to 9, while only one vase exists.
\newpage
\subsection{Environment 2}\label{App:Env2}
\begin{figure}[H]
    \centering
    \includegraphics[width=\linewidth, trim=7cm 5cm 5cm 2cm, clip]{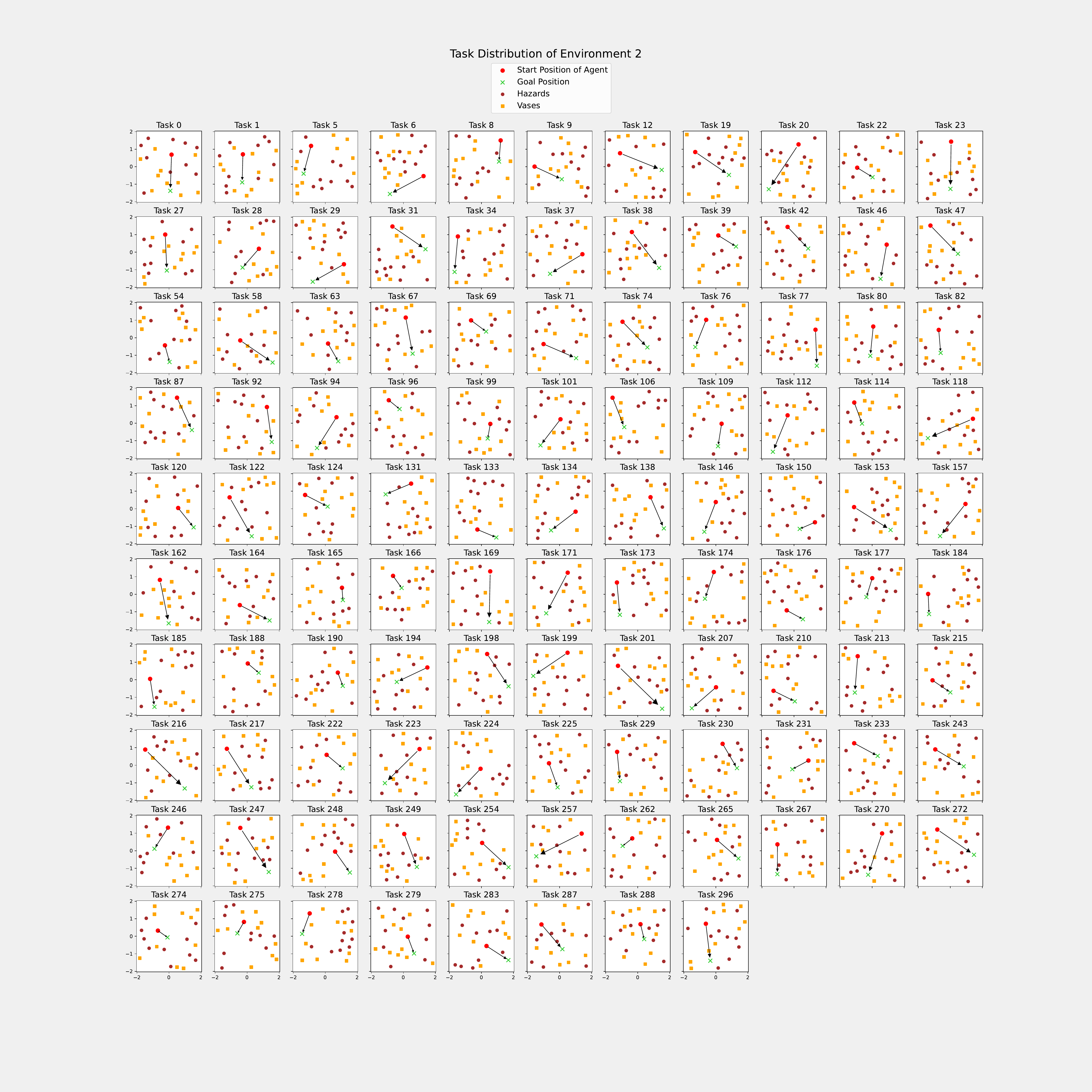}
    \caption{Illustrated are all 106 seeds comprising the task distribution within Environment 2. The tasks were uniformly drawn without replacement.}
    \label{fig:TaskDistrEnv2}
\end{figure}

Environment 2 consists of 106 tasks, illustrated in Figure \ref{fig:TaskDistrEnv2}, encompassing all seeds within the arc for the seed range of 0 to 300. The number of hazards and vases has now increased to ten each, obstructing a direct line of sight between the starting position and the goal in most cases due to several obstacles.

\newpage
\section{Fine-Tuning}\label{App:Rollouts}
This section clarifies the fine-tuning process for two specific tasks, namely task 122 and task 233, which are part of the broader task distribution denoted by \(T\). The fine-tuning process was meticulously examined using three distinct policy initializations to ensure a comprehensive understanding of the dynamics involved. The initializations are as follows:
\begin{enumerate}
    \item Random Initialization: This approach simulates the commencement of training from a foundational level, devoid of any prior learning or optimization.
    \item Pre-Optimized Policy Using TRPOLag: Here, the policy has been previously fine-tuned with the Trust Region Policy Optimization-Lagrangian (TRPOLag) algorithm, allowing us to assess the impact of prior optimization on the fine-tuning process.
    \item C-MAML Initialization with TRPOLag: This method involves initializing the policy through C-MAML, with TRPOLag being applied within the inner optimization loop, offering a unique perspective on the adaptability and efficiency of meta learning strategies in conjunction with TRPOLag.
\end{enumerate}
\begin{figure}
    \centering
    \begin{subfigure}{0.8\textwidth}
        \centering
        \includegraphics[trim = {0 0 0 1.8cm}, clip, width=\linewidth]{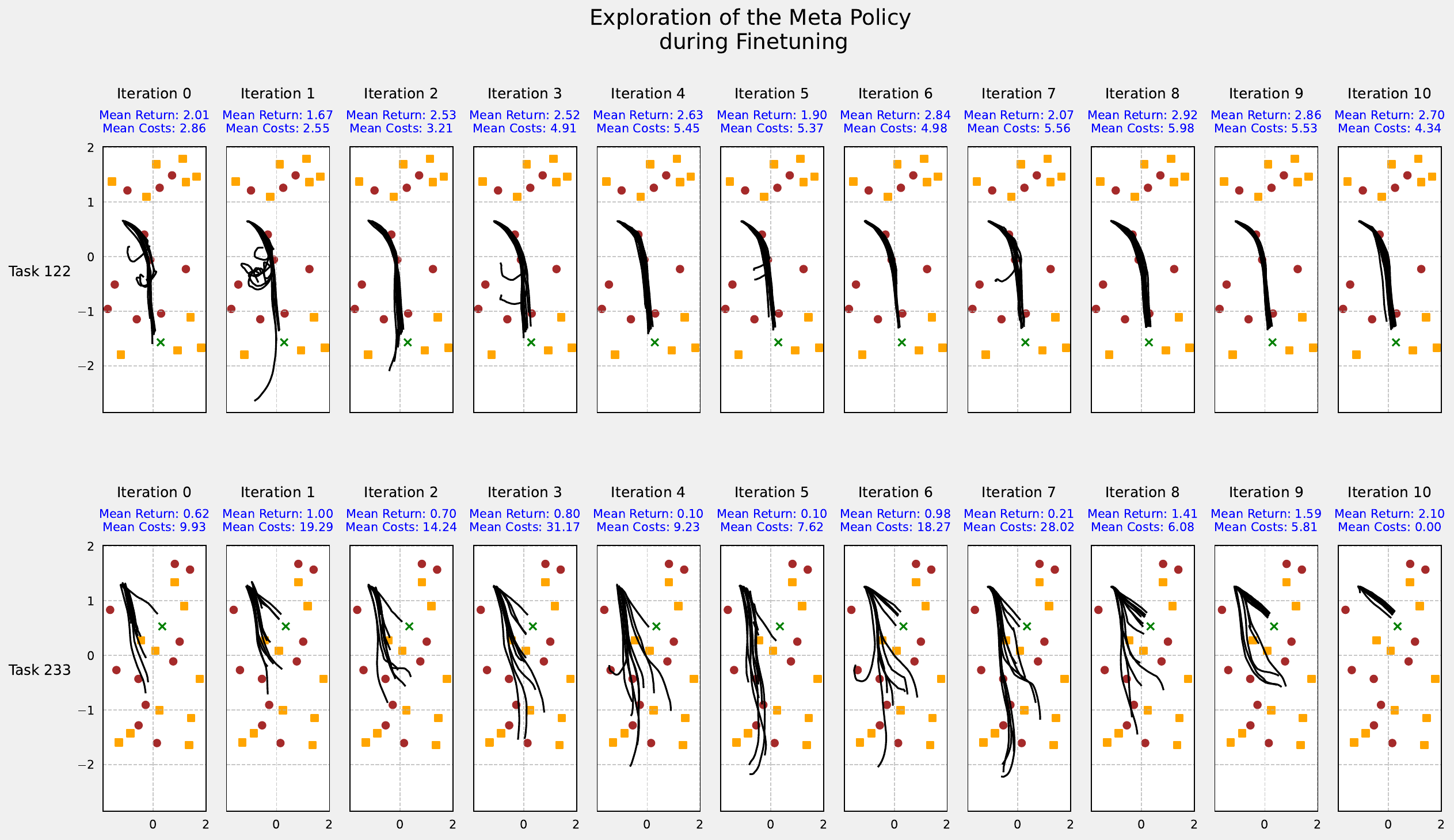}
        \caption{Rollout of the constrained meta-policy}
    \end{subfigure}
    \hfill
    \begin{subfigure}{0.8\textwidth}
        \centering
        \includegraphics[trim = {0 0 0 1.8cm}, clip, width=\linewidth]{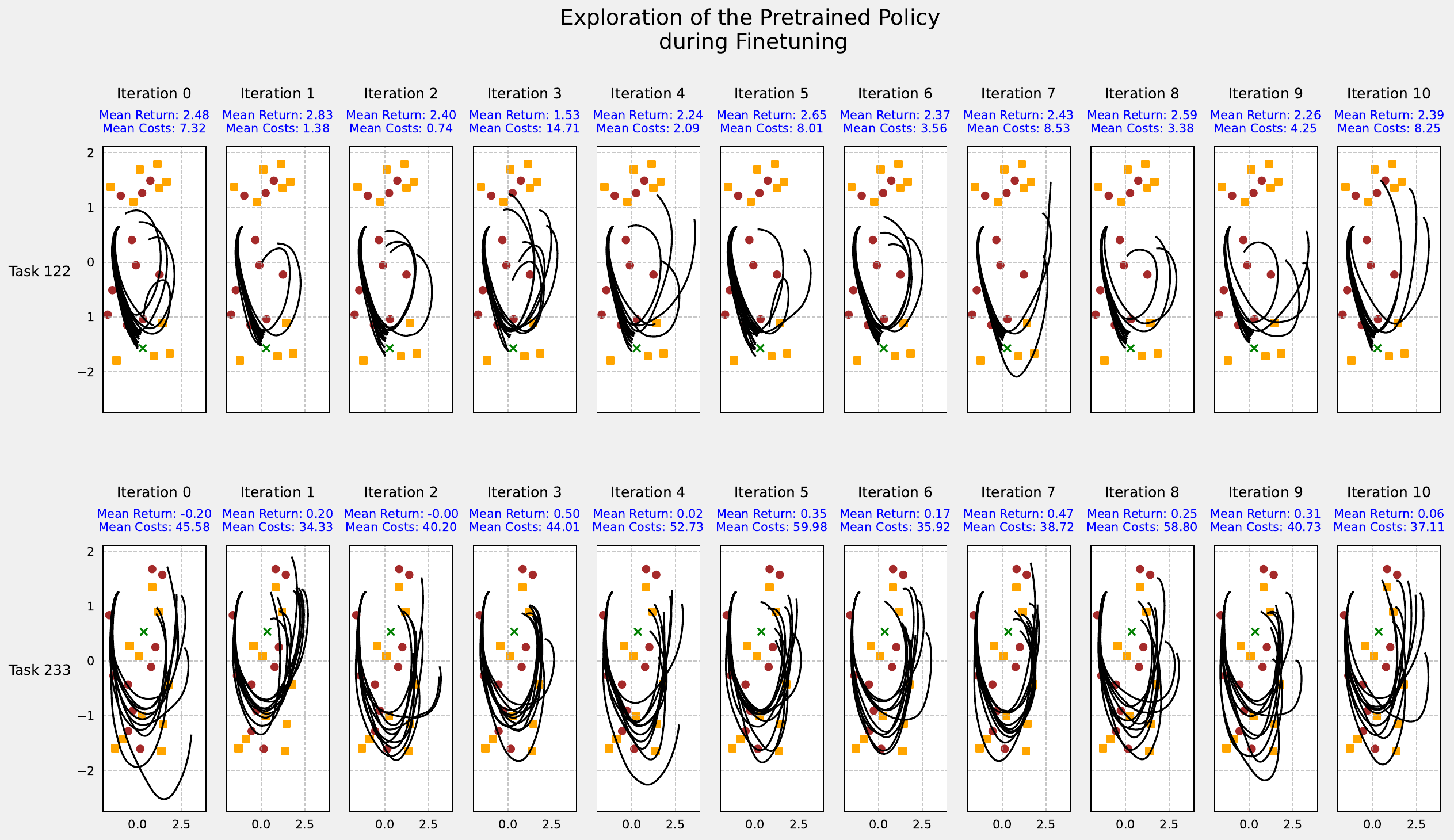}
        \caption{Rollout of a pretrained policy}
    \end{subfigure}
    \hfill
    \begin{subfigure}{0.8\textwidth}
        \centering
        \includegraphics[trim = {0 0 0 1.8cm}, clip, width=\linewidth]{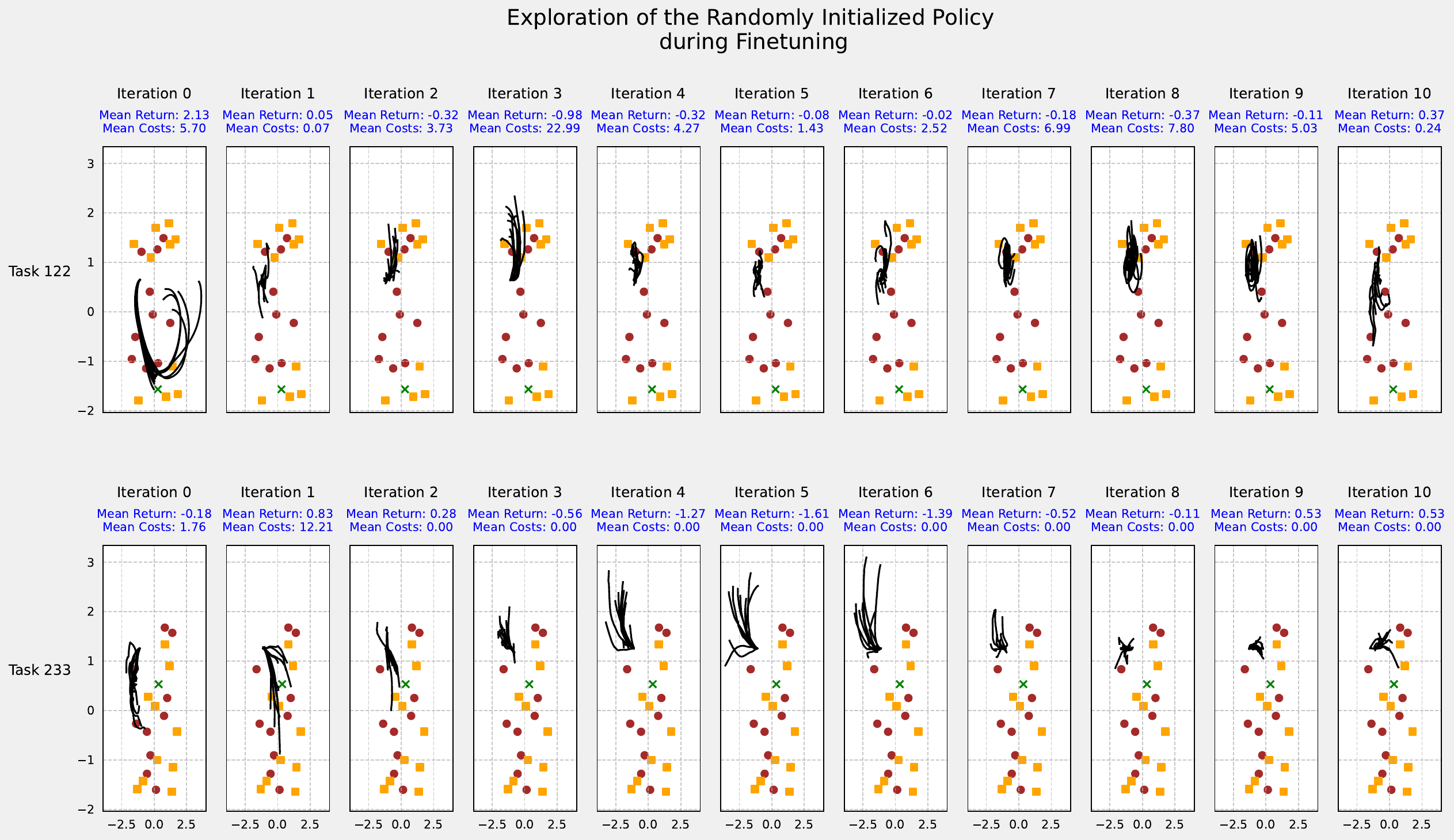}
        \caption{Rollout of a randomly initialized policy}
    \end{subfigure}
    \caption{The figure illustrates the behavior of a meta-policy trained with TRPOLag, a pretrained policy, and a randomly initialized policy during the fine-tuning process on two randomly selected tasks from the task distribution.}
    \label{fig:overall}
\end{figure}

\newpage
\section{Additional Results of C-MAML TRPOLag}\label{App:CMAMLTRPOLag}
\begin{figure}[H]
    \centering
    \includegraphics[width=\linewidth]{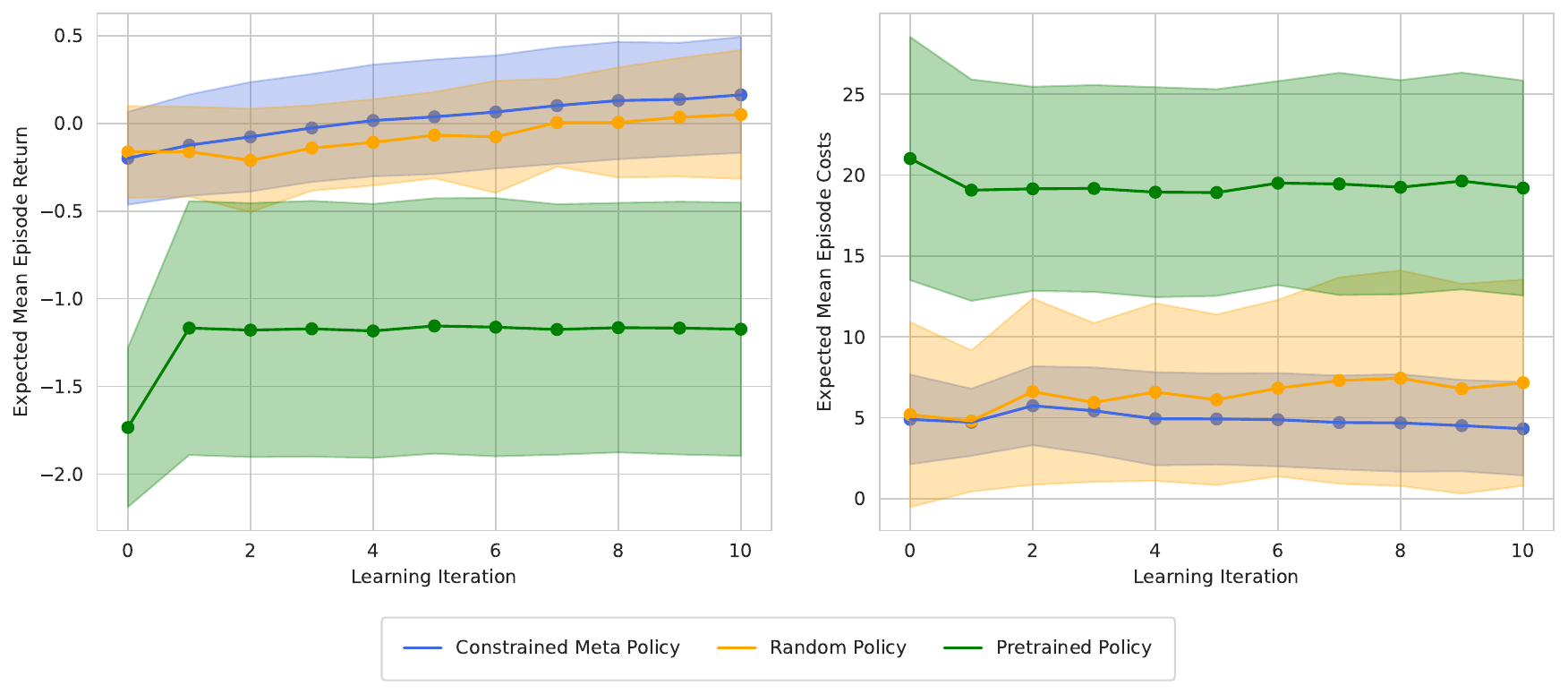}
    \caption{Progression of the mean episode returns and mean episode costs throughout the fine-tuning process. In contrast to Figure \ref{fig:TRPOLag_d5}, an average has been computed across the means of three different seeds, along with the indication of the mean standard deviation of these three seeds. The constrained meta-policy has undergone meta-training using TRPOLag. In fine-tuning, all policies have been adjusted with TRPOLag.}
    \label{fig:App_CMAMLTRPOLag}
\end{figure}

\section{Additional Results of C-MAML CPO}\label{App:CMAMLCPO}
\begin{figure}[H]
    \centering
    \includegraphics[width=\linewidth]{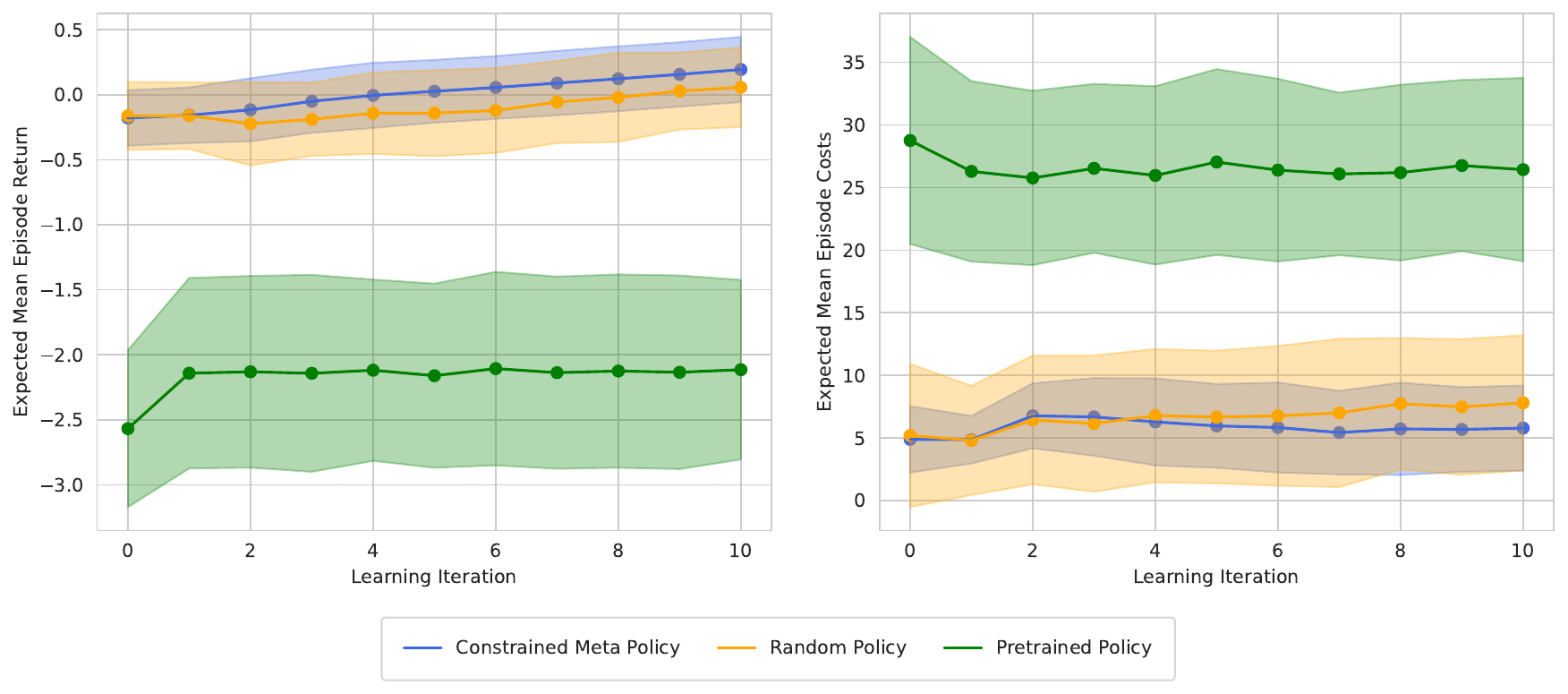}
    \caption{Progression of the mean episode returns and mean episode costs throughout the fine-tuning process. In contrast to Figure \ref{fig:CPO_d10}, an average has been computed across the means of three different seeds, along with the indication of the mean standard deviation of these three seeds. The constrained meta-policy has undergone meta-training using CPO. In fine-tuning, all policies have been adjusted with CPO.}
    \label{fig:App_CMAMLCPO}
\end{figure}

\newpage
\section{Impact of Difficulty Level}\label{App:DifficultyLevel}
\begin{figure}[H]
    \centering
    \begin{subfigure}{0.45\textwidth}
        \includegraphics[width=\linewidth]{figs/Env2-Task0.png}
        \caption{Task 0 of Environment 2}
        \label{fig:Env2_Task0}
    \end{subfigure}
    \hfill
    \begin{subfigure}{0.45\textwidth}
        \includegraphics[width=\linewidth]{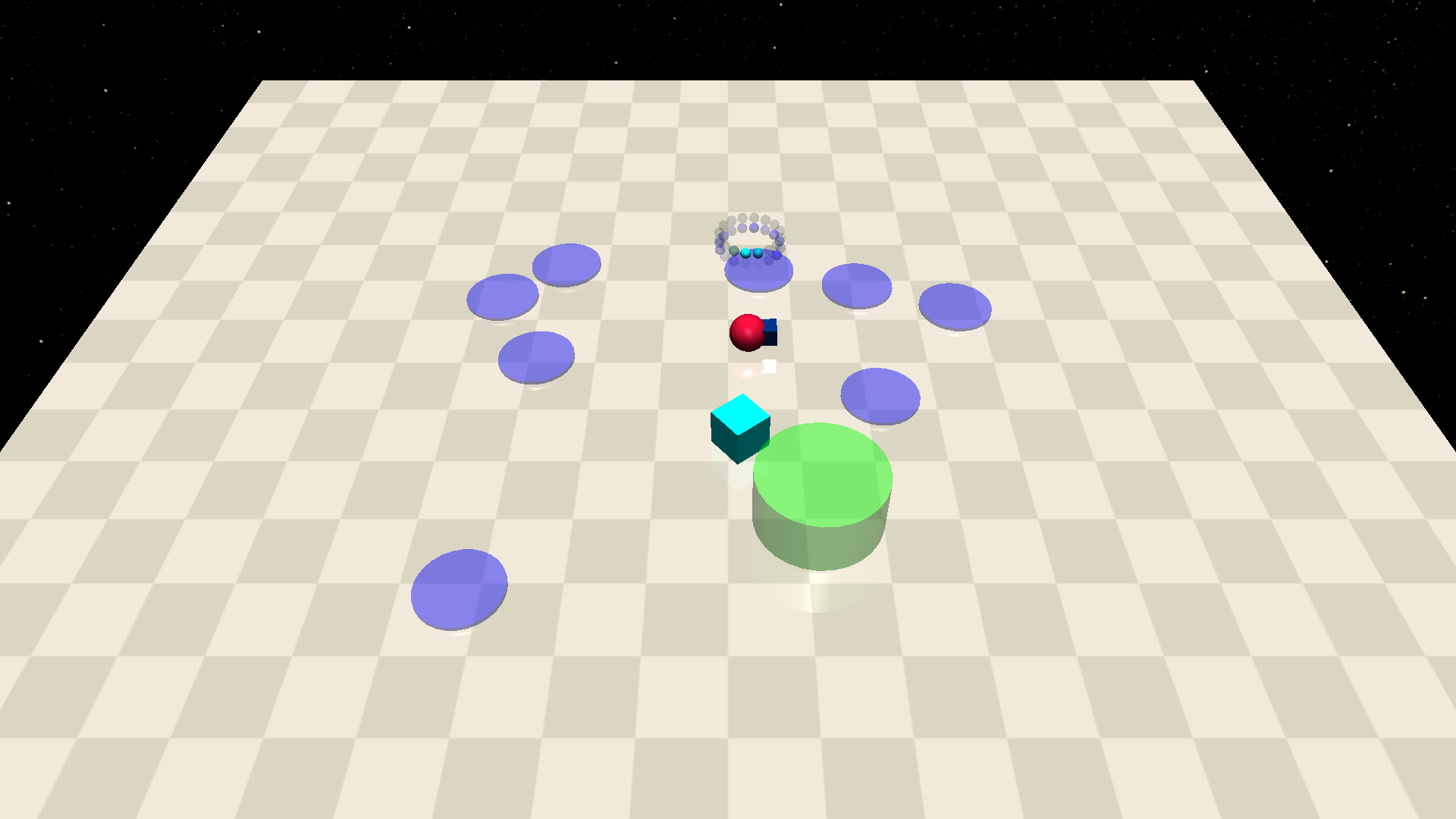}
        \caption{Task 0 of Environment 1}
        \label{fig:Env1_Task0}
    \end{subfigure}
    \caption{In the left illustration, Environment 2 consists of precisely ten hazards and ten vases in all tasks. Environment 1 mitigates the task's difficulty level by featuring only ten hazards and a single vase. Due to the reduced number of obstacles along the path from the agent to the goal, it is anticipated that the task becomes easier to solve.}
    \label{fig:Env_Comparison}
\end{figure}

\begin{figure}[H]
    \centering
    \includegraphics[width=\linewidth]{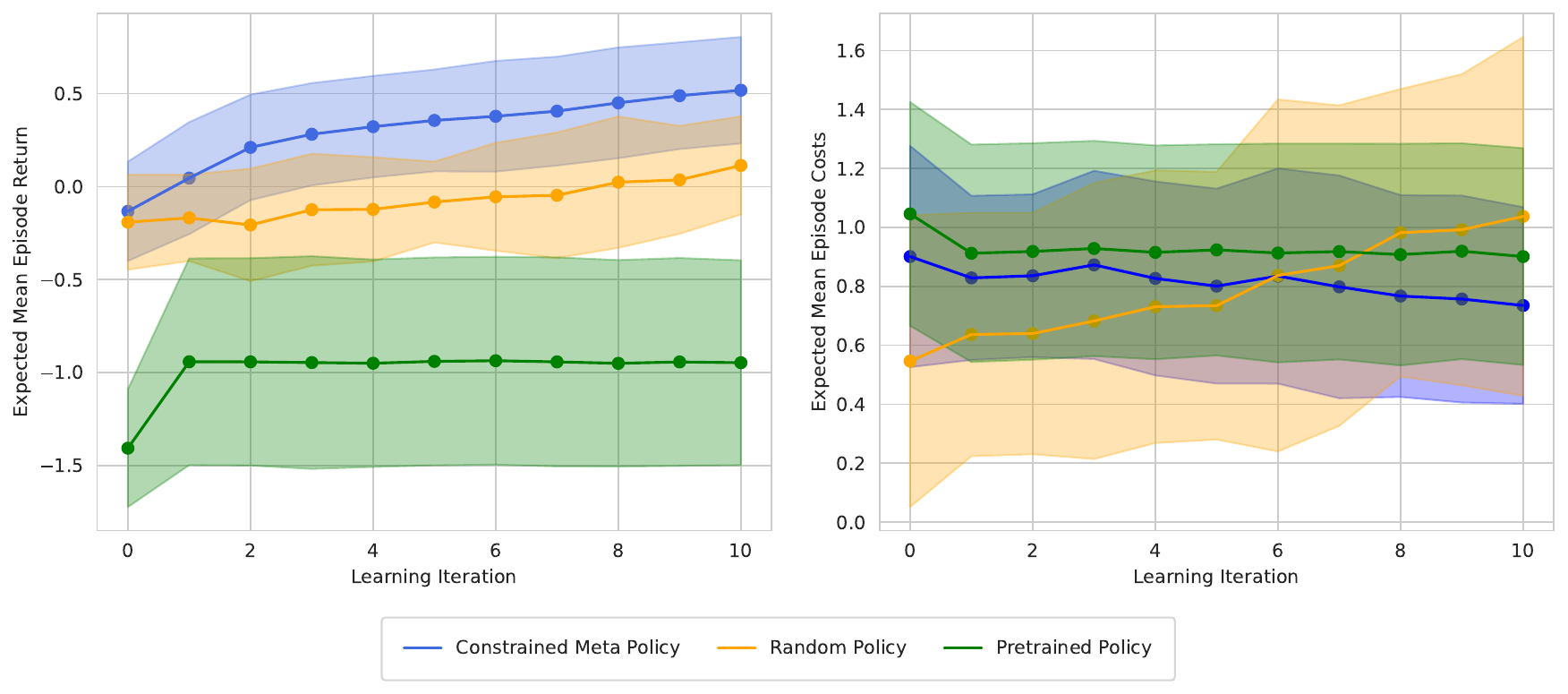}
    \caption{The trajectory of the mean episode return and mean episode costs during the course of fine-tuning on Environment 1 is depicted. As before, averaging was conducted across all tasks within the task distribution.}
    \label{fig:Influence_Difficulty}
\end{figure}

The transition from Environment 2 to Environment 1 allows for the modulation of difficulty levels, as the latter incorporates only ten obstacles, in contrast to the 20 obstacles depicted in Figure \ref{fig:Env_Comparison}. 
The reduction in environmental complexity is theoretically expected to facilitate the achievement of an agent's objectives while incurring lower episode costs. In order to assess the impact of varying difficulty levels on performance, we conducted additional experiments utilizing C-MAML with TRPOLag in Environment 1. The outcomes of these experiments are presented in Figure \ref{fig:Influence_Difficulty}.
The results reveal a notably quicker adaptation of the meta-policy compared to both the randomly initialized and pretrained policies. The pretrained policy fails to exhibit improvement after the initial adaptation step, maintaining a consistently low return level.

For all three policy types, the average episode costs exhibit a decrease during fine-tuning in both the meta-policy and the pretrained policy. After ten adaptation steps, the meta-policy reaches the cost limit of $d = 0.75$. In contrast, the random policy begins exploration during fine-tuning, resulting in a sharp increase in mean episode costs.

These results emphasize the robustness of the meta-policy in response to changes in the difficulty level of tasks and the environment.

\end{appendix}

\end{document}